%% file: paper.tex
\documentclass[11pt]{article}

\usepackage[preprint]{acl}

\usepackage{times}
\usepackage{latexsym}
\usepackage{amsmath,amssymb}
\input{math_commands.tex}

\usepackage{graphicx}
\graphicspath{{figures/}}
\usepackage{booktabs}
\usepackage{array}
\usepackage{multirow}
\usepackage{longtable}
\usepackage{algorithm}
\usepackage{algorithmic}
\usepackage{enumitem}
\usepackage{needspace}
\usepackage[most]{tcolorbox}
\usepackage[T1]{fontenc}
\usepackage[utf8]{inputenc}
\usepackage{microtype}
\usepackage{inconsolata}
\usepackage{pgf-pie}
\usepackage{float}
\usepackage{stfloats}
\definecolor{SkyBlue}{RGB}{108,166,205}
\hypersetup{
  pdftitle={Narrative Knowledge Weaver: Narrative-Centric Retrieval-Augmented Reasoning for Long-Form Text Understanding},
  pdfauthor={Qiuyu Tian, Fengyi Chen, Yiding Li, Youyong Kong, Fan Guo, Yuyao Li, Jinjing Shen, Zhijing Xie, Yiyun Luo, Xin Zhang, Yingce Xia, Zequn Liu}
}

\title{Narrative Knowledge Weaver: Narrative-Centric Retrieval-Augmented Reasoning for Long-Form Text Understanding}

\author{%
{\small
\begin{tabular}{@{}*{4}{>{\centering\arraybackslash}p{0.22\textwidth}}@{}}
Qiuyu Tian$^{1,2}$ & Fengyi Chen$^{3}$ & Yiding Li$^{5}$ & Youyong Kong$^{1}$ \\
Fan Guo$^{5}$ & Yuyao Li$^{5}$ & Jinjing Shen$^{5}$ & Zhijing Xie$^{5}$ \\
Yiyun Luo$^{5}$ & Xin Zhang$^{5}$ & Yingce Xia$^{2}$ & Zequn Liu$^{2}$\Thanks{Corresponding author.} \\[0.45em]
\multicolumn{4}{c}{%
\begin{tabular}{c}
\small $^{1}$Southeast University, Nanjing, China \\
\small $^{2}$Beijing Zhongguancun Academy, Beijing, China \\
\small $^{3}$Nanjing Normal University, Nanjing, China \\
\small $^{5}$ZhuiWen Technology Co., Ltd., Beijing, China
\end{tabular}}
\end{tabular}}}

\begin{document}

\maketitle

\begin{abstract}
Long-form narrative QA requires reasoning over evolving story worlds rather than isolated passages:
answers may depend on earlier goals, changing character states, social relations, causal triggers,
temporal position, and later consequences.  Existing retrieval and graph-augmented generation
methods improve evidence access, but their units--chunks, entities, relations, summaries, or tool
actions--do not directly encode how evidence functions in a story.  We introduce
\textbf{Narrative Knowledge Weaver} (NKW), a source-grounded framework that aligns textual evidence,
atomic facts, canonical graph structure, entity profiles, interactions, episodes, and storylines.
At query time, NKW uses text, graph, and narrative tools with post-retrieval reading skills to
assemble evidence and audit actor, scope, polarity, state, and temporal constraints.  Across STAGE,
FairytaleQA, and QuALITY, NKW is strongest on screenplay-level story-world QA while remaining
competitive on more passage-centered benchmarks.  Ablations, question-type analyses, graph-asset
statistics, and case studies show complementary benefits for character, scene, temporal, causal,
and narrative-progression reasoning.
\end{abstract}

\section{Introduction}

Long-form narrative understanding supports screenplay analysis, production workflows, literary
interpretation, and reading assistance.  Unlike ordinary long documents, narratives are organized
around scenes or chapters, recurring characters, changing states, and temporal-causal plot
progression.  Even with long-context LLMs, this remains challenging: the challenge is not only length,
but recovering how evidence functions inside an evolving story
world~\citep{graesser1990quest,kintsch1988construction,zwaan1998situation,liu2024lost}.

Retrieval-augmented generation (RAG) partially addresses this challenge by grounding model
responses in external evidence~\citep{lewis2020rag,izacard-grave-2021-fid}.  This makes
it useful for locating relevant passages and reducing unsupported generation in long documents.
Graph-based RAG
organize retrieved content through entity, relation, or community-level abstractions
\citep{edge2024fromlocal,guo-etal-2025-lightrag,gutierrez2024hipporag,zhu2025kg2rag}, while
tool- and retrieval-control RAG methods adapt the retrieval process during generation
\citep{jiang2023flare,trivedi2023ircot,asai2023selfrag,du2026arag}.  However, these methods are largely developed
for factual, open-domain, or general long-document settings, whereas narratives require different
evidence organization.

We identify two mismatches between general-purpose RAG and narrative QA.  First, narrative evidence
is functional, not only locational: the same passage may serve as premise, turning point,
consequence, state update, or distractor.  For a why-question, retrieval may find the decision scene
while missing the earlier motivating interaction.  Chunks, entities, relations, and community
summaries locate material, but do not encode its story role.  Second, characters and relations are
dynamic.  Narrative questions ask what a character believes, wants, knows, or feels at a plot
moment, or how attitudes change after a scene, requiring evolving states and relationships rather
than a flat entity profile. Together, these mismatches make it difficult for general-purpose RAG
systems to assemble evidence that is not only relevant in content, but also appropriate in narrative
function and dynamic character context.

To this end, We present \textbf{Narrative Knowledge Weaver} (NKW), a narrative-centric RAG system for these two
mismatches.  To represent functional evidence roles, NKW extracts source-grounded
narrative assets and aggregates events, interactions, and occasions into higher-level episodes and
storylines, so that evidence can be organized by its contribution to plot progression rather than
by passage similarity alone. To capture dynamic characters and relations, NKW separates stable identity from changing state.
It first links names, aliases,
pronouns, and scene-specific references to the stable narrative entity. It then uses
source-grounded atomic facts to build time-sensitive character profiles.
Thus, characters are represented not as static graph nodes, but as evolving narrative states
anchored in textual evidence. Furthermore, NKW
provides a channel-separated tool interface over text, graph, and narrative views, together with
post-retrieval reading skills that assemble and audit evidence for actor, scope, polarity, and
temporal position at inference time.

We evaluate NKW on on three long-form narrative QA datasets, STAGE~\citep{tian2026stage}, FairytaleQA~\citep{xu2022fairytaleqa}, and
QuALITY~\citep{pang2022quality}, across seven LLM backbones against Hybrid RAG, GraphRAG, LightRAG,
HippoRAG, and A-RAG.  Results show the largest benefits when questions require reasoning over
evolving states, relations, temporal order, causal motivation, and plot progression.  Ablations,
STAGE question-type breakdowns, graph-asset statistics, case studies, and two downstream
applications further show when the narrative structure contributes.

Our key contributions are:
\begin{itemize}[leftmargin=*,nosep,topsep=2pt,partopsep=0pt]
    \item \textbf{A narrative-centric RAG system.} NKW represents long-form narratives through
    canonical entity graphs, source-grounded narrative assets, episode/storyline aggregation, and
    atomic-fact-based character profiling.
    \item \textbf{Inference tools for narrative evidence assembly.} NKW uses channel-separated
    access to text, graph, and narrative views, with post-retrieval reading skills to audit actor,
    scope, polarity, and temporal position.
    \item \textbf{Comprehensive evaluation and applications.} We evaluate NKW on STAGE,
    FairytaleQA, and QuALITY across seven LLM backbones, with ablations, question-type analyses,
    case studies, and two downstream narrative applications.
\end{itemize}

\section{Related Work}

\subsection{Narrative and Long-Form Text Understanding}

Narrative understanding requires reasoning over evidence distributed across an evolving story
world.  QuALITY~\citep{pang2022quality}, FairytaleQA~\citep{xu2022fairytaleqa}, and
STAGE~\citep{tian2026stage} cover long-passage reading, fine-grained story questions, and
full-screenplay story-world reasoning.  These benchmarks test final answers, but they provide
limited supervision for the intermediate story representation needed to recover those answers.

This representation gap is also reflected in prior work on event
causality~\citep{sun2024eventcausality} and event grounding~\citep{li2024eventground}, which
emphasizes connecting what happens, why it happens, and where it is supported.  Discourse
comprehension~\citep{graesser1994constructing}, construction-integration
theory~\citep{kintsch1988construction}, and situation-model theory~\citep{zwaan1998situation}
similarly treat narrative meaning as evolving states, goals, causal links, and perspectives.  NKW
operationalizes this view as source-grounded layers for facts, character states, interactions,
episodes, and storylines.

\subsection{Graph Construction and Graph-Augmented Retrieval}

Graph construction supplies useful backbone machinery but is not the main target of NKW.
Document-level IE and coreference methods model entities, relations, events, and normalized
mentions~\citep{yao-etal-2019-docred,wadden-etal-2019-dygiepp,lin-etal-2020-oneie,barhom-etal-2019-revisiting,cattan-etal-2020-benchmarking};
LLM-based methods such as EDC~\citep{zhang2024edc} and Docs2KG~\citep{sun2024docs2kg} extend this
pipeline to flexible documents.  These works primarily target extraction coverage, whereas NKW uses
a stable graph backbone while making events, atomic facts, entity profiles, interactions, episodes,
storylines, and provenance explicit narrative assets.

Graph-augmented RAG and graph-grounded reasoning expose structured evidence for retrieval and
multi-step inference, including GraphRAG~\citep{edge2024fromlocal},
LightRAG~\citep{guo-etal-2025-lightrag}, HippoRAG~\citep{gutierrez2024hipporag},
KG2RAG~\citep{zhu2025kg2rag}, A-RAG~\citep{du2026arag}, RoG~\citep{luo2024rog},
ToG~\citep{sun2024tog}, and KG-Agent~\citep{jiang2025kgagent}.  NKW connects to this line but
organizes retrieval around narrative dynamics: local facts, character profiles, interactions, event
traces, episodes, and storylines.

\section{The Narrative Knowledge Weaver Framework}
\label{sec:framework}

We propose \textbf{Narrative Knowledge Weaver}, a framework for narrative-centric knowledge
modeling and reasoning over long-form text.  Let \(D=\{c_i\}_{i=1}^{n}\) denote the source chunks.
NKW constructs a source-grounded asset bundle
\(\mathcal{B}=(G,\mathcal{U},\mathcal{F},\mathcal{P},\mathcal{H},\mathcal{X})\), where
\(G=(V,R)\) is a canonical entity--relation graph, \(\mathcal{U}\) contains events, interactions,
and occasions, \(\mathcal{F}\) contains atomic facts, \(\mathcal{P}\) contains entity profiles,
\(\mathcal{H}\) contains episode/storyline structures, and \(\mathcal{X}\) indexes links back to
source evidence.  The system separates a construction-time agent that builds this bundle from a
query-time reasoning agent that uses it with source evidence to answer questions.

\begin{figure*}[t]
    \centering
    \includegraphics[width=\linewidth]{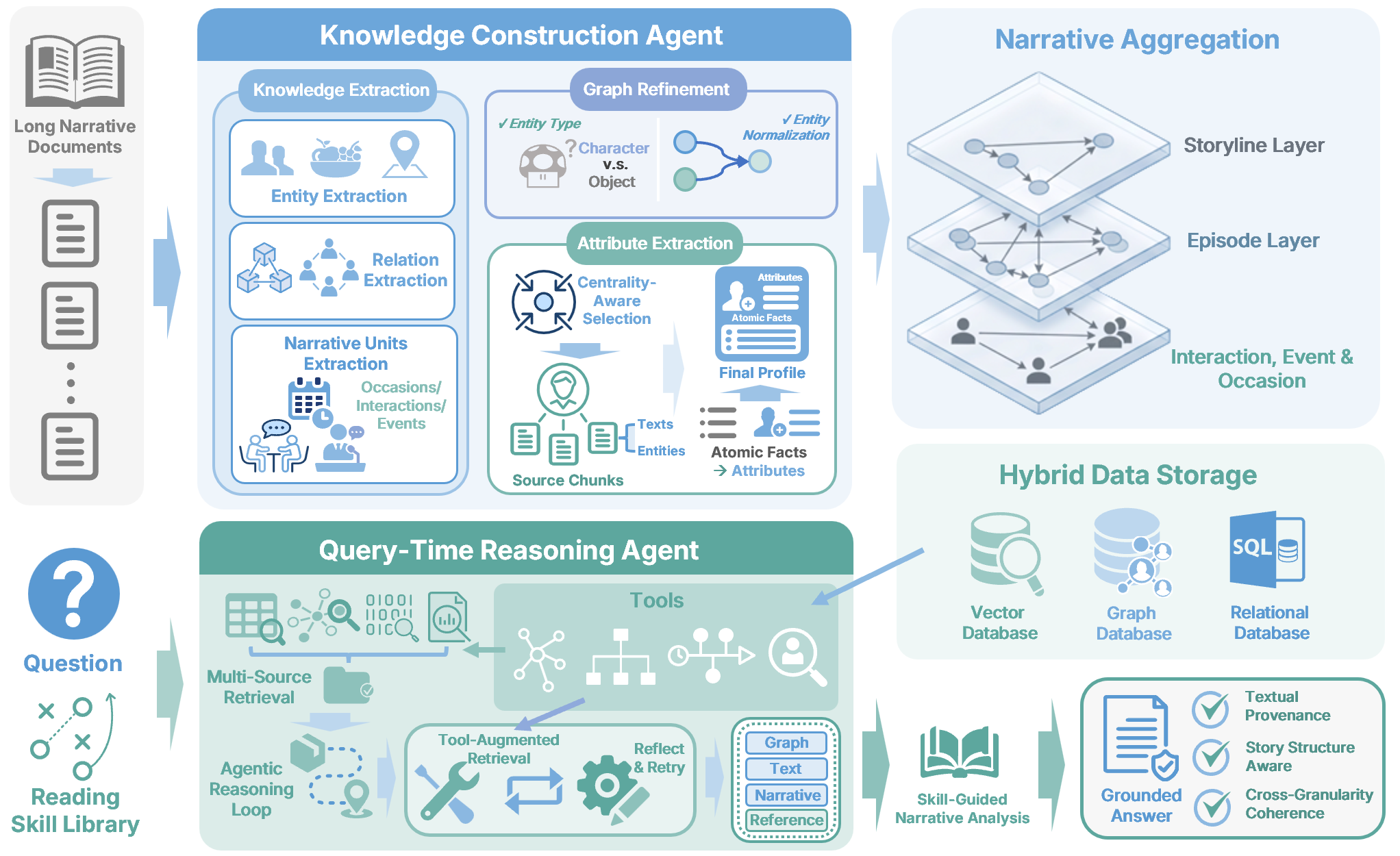}
    \caption{\textbf{Narrative-centric framework overview.} The framework separates
    construction-time graph building from query-time reasoning.}
    \label{fig:overall-architecture}
\end{figure*}

\subsection{Construction-Time Asset Building}
\label{sec:asset-building}
The construction-time agent builds complementary source-grounded assets in a fixed order: a stable
entity--relation graph backbone, narrative units and local evidence records, and then
canonicalized entity profiles and graph-derived facts.

\subsubsection{Canonical Entity--Relation Graph Backbone}
\label{sec:graph-backbone}
NKW first builds a canonical graph backbone over stable narrative entities.  Each chunk \(c_i\) is
processed by an entity--relation extractor that emits entity rows
\(v=(\mathrm{name},\mathrm{type},\mathrm{desc},S_v)\) and relation rows
\(r=(u,v,\mathrm{desc},\mathrm{keywords},w,S_r)\), where \(S_v\) and \(S_r\) are source identifiers.
The native entity schema is restricted to stable story referents, including
characters, groups, locations, time points, objects, institutions, social roles, and concepts.
Narrative process units such as events, interactions, occasions, episodes, storylines, and
scene/chapter titles are not treated as native entity types in this backbone.

Relations in this backbone are binary edges between stable entities.  They are stored as unordered
entity pairs by default, with relation keywords used as topical descriptors rather than as a fixed
relation ontology.  N-ary statements are decomposed into binary edges, and missing endpoints are
materialized as lightweight proxy entities so that the relation can still be source-grounded.
During merge and upsert, same-name entities and same-pair relations are consolidated, long
descriptions are compacted, and graph, vector, entity--chunk, and relation--chunk indexes are kept
in sync.  This backbone provides stable entity and relation grounding for later narrative assets.

\subsubsection{Source-Grounded Narrative Asset Extraction}
\label{sec:narrative-assets}
In parallel with the graph backbone, NKW extracts source-grounded narrative assets from each chunk.
The base narrative units are \emph{events}, which describe actions, discoveries, decisions, arrivals,
departures, physical processes, or state changes; \emph{interactions}, which describe directed
communication, conflict, cooperation, or influence between participants; and \emph{occasions}, which
describe stable situations, institutional constraints, social norms, background conditions, or
scene-like contexts.  The extractor also produces source-grounded atomic facts and entity attributes.

Each narrative unit remains linked to its source evidence.  Interactions store structured metadata
such as subject, object, interaction type, related event, and related occasion; events and occasions
store participants, settings, temporal cues, and local context when available.  These structured
fields support later canonicalization and retrieval, while visible narrative content is rendered as
clean natural language rather than as internal identifiers or storage metadata.

\subsubsection{Entity Canonicalization and Evidence Enrichment}
\label{sec:graph-refinement}
After merging the graph backbone, NKW canonicalizes salient mentions so that aliases and local
descriptions of the same character, object, place, institution, or plan share one referent.  The
canonical map is applied to structured narrative fields--participants, interaction subject/object
fields, atomic-fact subject/object fields, and owners of entity attributes--while natural-language
evidence remains unchanged.

Only occasions are injected into the main graph, as contextual nodes linked to relevant participants
or settings.  Events and interactions stay in the narrative store and vector index for retrieval and
episode/storyline construction, preserving source-grounded evidence without destabilizing the
entity-relation graph.  For entities with graph degree at least two in our experiments, NKW revisits
linked source chunks, extracts entity-specific atomic facts about states, actions, goals, causal
roles, mental states, and persistent relations, and derives a compact attribute dictionary and
summary.  Duplicate facts and overlapping attributes are merged, then stored in a semantic index for
character, relation, and state reasoning.

\subsection{Global Episode and Storyline Construction}
\label{sec:narrative-hierarchy}
Local entity aggregation is not sufficient for questions about plot development, motivation, or
long-range consequence.  NKW therefore aggregates local narrative units
\(\mathcal{U}=\mathcal{E}\cup\mathcal{I}\cup\mathcal{O}\) into episodes using a coherence function
\(\phi(u,z)\) over shared participants, goals, conflicts, occasions, temporal continuity, causal
dependency, semantic similarity, and source support:
\begin{equation}
\mathcal{Z}=\operatorname{Cluster}(\mathcal{U};\phi),\quad
\widehat{G}_{\mathcal{Z}}=\operatorname{SABER}(\mathcal{Z},R_{\mathcal{Z}}).
\end{equation}
Local episodes may retain document, scene, or chunk provenance, but they are not treated as
storyline boundaries.

Storyline construction is global over the workspace.  NKW predicts episode relations
\(R_{\mathcal{Z}}\subseteq \mathcal{Z}\times\{\emph{causes},\emph{elaborates},\emph{precedes}\}\times
\mathcal{Z}\) using source order, shared participants, lexical similarity, shared graph neighbors,
and episode-content similarity.  SABER cleans the noisy episode graph into the DAG
\(\widehat{G}_{\mathcal{Z}}\): strongly connected components are broken by removing weak edges, and
only after the DAG skeleton is stable are ambiguous triangle shortcuts sent for LLM adjudication.
Storylines \(\mathcal{L}\) are then extracted as episode chains through a trunk--branch view, with
relations such as blocking, resolving, prerequisite, conflict, and parallel arcs predicted over
\(\mathcal{L}\).  Thus, narrative units support episodes, episodes support storylines, and
storylines provide compact handles for long-range temporal and causal reasoning.
Appendix~\ref{app:narrative-aggregation-details} details the aggregation procedure.

\subsection{Narrative-Grounded Retrieval and Reasoning}
\label{sec:storage-application}

\subsubsection{Evidence Views}
The constructed asset bundle provides complementary evidence views over the same narrative:
source passages for recall, atomic facts for compact local propositions, entity profiles for
states and relations, event/interaction/occasion records for local structure, and episode/storyline
layers for long-range progression.  Each view remains linked to source evidence.  When an entity
profile contains many supporting atomic facts, query-time retrieval does not expose the full fact
list by default.  Instead, for an entity \(e\) and question \(q\), NKW returns
\(\operatorname{TopK}_{k=5}\{f\in\mathcal{F}(e):\operatorname{sim}(q,f)\}\) in our experiments.
This prevents salient
entities from overwhelming the evidence packet while preserving fine-grained source grounding.

\subsubsection{Query-Time Tools}
NKW exposes these views through a compact tool interface instead of a single flat retriever.  The
interface provides text access for source evidence, graph access for entity and relation grounding,
and narrative access for temporal, causal, and storyline progression.  These tools add no external
knowledge; they only route queries over source-grounded assets, allowing the agent to combine local
text, graph structure, and narrative structure while preserving provenance.
Appendix~\ref{app:tools-implementation} provides the detailed tool taxonomy used in inference.

At query time, retrieval is formulated as evidence assembly.  Given tools \(\mathcal{T}\), the agent
decomposes a question \(q\) into requirements over local facts, entity states, interactions,
temporal order, causal explanation, or plot progression, then builds an evidence packet
\(\mathcal{E}_q=\operatorname{Assemble}(q,\mathcal{B},\mathcal{T})\).  The final answer is generated
from \(\mathcal{E}_q\): structured signals guide navigation and constraint, while textual evidence
remains the basis for answer support.
Appendix~\ref{app:qa-procedure} gives the end-to-end QA procedure.

\subsubsection{Post-Retrieval Reading Skills}
Post-retrieval \emph{reading skills} further guide interpretation of the assembled evidence.  They
are implemented as lightweight Reading Skill Cards: final-stage operators seeded by cognitive QA and
narrative-comprehension theories, then calibrated to benchmark question patterns.  They do not
change the constructed graph or retrieve new evidence; instead, they audit how an existing evidence
packet supports actor, scope, time, state, causal, discourse, and option-level constraints.
Operationally, a selector injects one to three short card operators into the final-answer prompt to
reduce wrong-actor, wrong-time, wrong-cause, wrong-state, and wrong-scope readings.
Appendix~\ref{app:reading-skills} summarizes these cards, and Appendix~\ref{app:mcq-option-admission}
details the multiple-choice option-admission procedure.

\section{Experimental Setup}
\label{sec:setup}

\paragraph{Datasets.}
We evaluate on three narrative and long-form QA benchmarks.  \textbf{STAGE}~\citep{tian2026stage}
contains 151 movie screenplays and 5,010 questions, with an average document length of 27,546 in
our processed benchmark copy.  The 151 screenplays include 109 English and 42 Chinese scripts, and
the questions are annotated with six narrative reasoning types covering character, scene, temporal,
causal, relation, and progression reasoning.  \textbf{QuALITY}~\citep{pang2022quality} contains 162
long English passages and 2,523 multiple-choice questions, with an average length of 4,115.
\textbf{FairytaleQA}~\citep{xu2022fairytaleqa} contains 278 children's stories and 10,580
expert-authored questions, with an average length of 2,165.  Document length is measured as
whitespace-token word count for English text and character count for Chinese text in the processed
benchmark copy.  Appendix~\ref{app:stage-question-type} reports the STAGE type-level
distribution and performance breakdown used in our analysis.

\paragraph{Baselines.}
We choose baselines to cover two axes central to our comparison: graph structure and tool-mediated
retrieval. \textbf{Hybrid RAG} represents strong non-graph retrieval. \textbf{GraphRAG},
\textbf{LightRAG}~\citep{guo-etal-2025-lightrag}, \textbf{HippoRAG}~\citep{gutierrez2024hipporag},
and \textbf{A-RAG}~\citep{du2026arag} cover recent graph- and tool-augmented RAG paradigms,
including community summaries, entity-relation indexing, associative graph activation, and
agent-controlled retrieval interfaces. \textbf{NKW} tests their combination through text, graph,
and narrative tools over narrative-centric graph structures
(Appendix~\ref{sec:tools-and-usage}).

\vspace{-4pt}
\begin{itemize}[leftmargin=2em, itemsep=2pt, topsep=2pt]
    \item All baselines use a \textbf{600-token chunk size}.
    \item Model-family comparisons use non-thinking Qwen3 and Llama-3.1 backbones, plus GPT-5.5 as a closed-source reference.  Table names omit checkpoint suffixes for readability; Qwen3-30B refers to \textbf{Qwen3-30B-A3B-Instruct-2507}, Qwen3-235B refers to \textbf{Qwen3-235B-A22B-FP8}, and Llama rows use the corresponding Llama-3.1 Instruct checkpoints.
    \item Component ablations are run only with \textbf{Qwen3-235B}.
\end{itemize}
\vspace{-4pt}

\paragraph{Evaluation Metrics.}
For STAGE and semantic-answer FairytaleQA items, we use a DeepSeek-V4 LLM-based evaluator that
judges each system answer against the question and ground-truth answer.  QuALITY is a
multiple-choice benchmark, so correctness is determined by whether the selected option matches the
gold option.  Across tables, \textbf{Overall} denotes the benchmark's primary correctness metric,
and \textbf{Pass@5} denotes whether any of five sampled answers is correct.

\paragraph{Ablation Protocol.}
Component ablations use the same Qwen3-235B backbone, decoding configuration, question sets, and
benchmark-specific metrics as the full system.  We remove one structural component at a time while
holding the query-time reasoning budget fixed, isolating gains from narrative-centric structure
rather than context size or sampling budget.  For STAGE, we further report type-level effects over
character, scene, temporal, causal, relation, and progression questions.

\section{Experimental Results and Analysis}
\label{sec:results-analysis}

\subsection{Question Answering}
\label{sec:qa-results}

\paragraph{Main results.}
Table~\ref{tab:stage-main-results} reports the STAGE result block in the same
backbone-by-access-structure format used by STAGE. We report it first because it is the most direct
story-world benchmark, while treating all benchmarks as one narrative suite.

\begin{table*}[t]
\centering
\scriptsize
\caption{Main QA results on STAGE across backbones. Bold marks the best method separately for Overall and Pass@5 within each backbone row.}
\label{tab:stage-main-results}
\setlength{\tabcolsep}{1.25pt}
\resizebox{\textwidth}{!}{%
\begin{tabular}{l|cc|cc|cc|cc|cc|cc}
\toprule
\multirow{2}{*}{\textbf{Backbone}} &
\multicolumn{2}{c|}{\textbf{Hybrid RAG}} &
\multicolumn{2}{c|}{\textbf{GraphRAG}} &
\multicolumn{2}{c|}{\textbf{LightRAG}} &
\multicolumn{2}{c|}{\textbf{HippoRAG}} &
\multicolumn{2}{c|}{\textbf{A-RAG}} &
\multicolumn{2}{c}{\textbf{NKW (Ours)}} \\
& \textbf{Overall} & \textbf{Pass@5}
& \textbf{Overall} & \textbf{Pass@5}
& \textbf{Overall} & \textbf{Pass@5}
& \textbf{Overall} & \textbf{Pass@5}
& \textbf{Overall} & \textbf{Pass@5}
& \textbf{Overall} & \textbf{Pass@5} \\
\midrule
Qwen3-8B & \textbf{0.4216} & 0.4637 & 0.2854 & 0.3124 & 0.3685 & 0.4329 & 0.2643 & 0.2890 & 0.2315 & 0.2741 & 0.3852& \textbf{0.5014}\\
Qwen3-30B & 0.5144 & 0.5563 & 0.3028& 0.4126 & 0.3874 & 0.5393 & 0.3186 & 0.3441 & 0.2980 & 0.3756 & \textbf{0.5892} & \textbf{0.6643} \\
Qwen3-235B & 0.6056& 0.6465& 0.3206& 0.4790& 0.4407 & 0.5988 & 0.4014 & 0.4301 & 0.3625 & 0.4862 & \textbf{0.7012} & \textbf{0.8148} \\
Llama-3.1-8B & \textbf{0.3854} & 0.4124 & 0.2916 & 0.3246 & 0.3419 & 0.4106 & 0.2719 & 0.2932 & 0.2146 & 0.2529 & 0.3721 & \textbf{0.4854} \\
Llama-3.1-70B & 0.5124 & 0.5547 & 0.3122& 0.4691& 0.3912 & 0.5216 & 0.4102 & 0.4415 & 0.3349 & 0.4214 & \textbf{0.5629} & \textbf{0.6431} \\
Llama-3.1-405B & 0.6112& 0.6385& 0.3525& 0.4982& 0.4527 & 0.5894 & 0.4874 & 0.5112 & 0.3952 & 0.5020 & \textbf{0.6535} & \textbf{0.7429} \\
GPT-5.5 & 0.6216 & 0.6525& 0.4024& 0.5357& 0.4784 & 0.6122 & 0.5118 & 0.5429 & 0.4216 & 0.5325 & \textbf{0.6984} & \textbf{0.8114} \\
\bottomrule
\end{tabular}
}
\end{table*}

Across STAGE backbones, NKW obtains the best Pass@5 in every row and the best Overall score for all
medium, large, and closed-source backbones.  The main exceptions are the two smallest open-source
backbones, where Hybrid RAG gives slightly higher one-shot Overall accuracy for Qwen3-8B and
Llama-3.1-8B, while NKW still improves Pass@5.  This suggests that structured narrative evidence is
most useful when the backbone can reliably use multi-source evidence, and that repeated sampling
reveals gains from the richer evidence space even when one-shot selection is less stable.

Appendix~\ref{app:stage-question-type} further breaks down the STAGE gains by six question types,
and Appendix~\ref{app:stage-case-analysis} gives success and failure cases showing where NKW
resolves baseline errors and where it can still confuse local triggers with salient downstream
events.

\paragraph{Secondary benchmarks.}
Tables~\ref{tab:fairytale-results} and~\ref{tab:quality-results} record the supporting
results on FairytaleQA and QuALITY, which complement STAGE by testing narrative
comprehension over shorter stories and long-form multiple-choice passages.

\begin{table*}[t]
\centering
\scriptsize
\caption{Supporting QA results on FairytaleQA. Bold marks the best method separately for Overall and Pass@5 within each backbone row.}
\label{tab:fairytale-results}
\setlength{\tabcolsep}{1.25pt}
\resizebox{\textwidth}{!}{%
\begin{tabular}{l|cc|cc|cc|cc|cc|cc}
\toprule
\multirow{2}{*}{\textbf{Backbone}} &
\multicolumn{2}{c|}{\textbf{Hybrid RAG}} &
\multicolumn{2}{c|}{\textbf{GraphRAG}} &
\multicolumn{2}{c|}{\textbf{LightRAG}} &
\multicolumn{2}{c|}{\textbf{HippoRAG}} &
\multicolumn{2}{c|}{\textbf{A-RAG}} &
\multicolumn{2}{c}{\textbf{NKW (Ours)}} \\
& \textbf{Overall} & \textbf{Pass@5}
& \textbf{Overall} & \textbf{Pass@5}
& \textbf{Overall} & \textbf{Pass@5}
& \textbf{Overall} & \textbf{Pass@5}
& \textbf{Overall} & \textbf{Pass@5}
& \textbf{Overall} & \textbf{Pass@5} \\
\midrule
Qwen3-8B & 0.7324 & 0.7457 & 0.3121 & 0.3650 & 0.6792 & \textbf{0.8229} & \textbf{0.7708} & 0.8131 & 0.3873 & 0.4726 & 0.6525& 0.8015\\
Qwen3-30B & 0.8115 & 0.8239 & 0.3953 & 0.4626 & \textbf{0.8563} & 0.9241 & 0.8012 & 0.8124 & 0.4699 & 0.5302 & 0.8515 & \textbf{0.9452} \\
Qwen3-235B & 0.8406 & 0.8502 & 0.4487 & 0.5254 & \textbf{0.8752} & 0.9334 & 0.8256 & 0.8393 & 0.5871 & 0.7029 & \textbf{0.8752} & \textbf{0.9545} \\
Llama-3.1-8B & \textbf{0.7216} & 0.7314 & 0.3340 & 0.3898 & 0.7155 & \textbf{0.7723}& 0.7082 & 0.7197 & 0.4250 & 0.5211 & 0.6842& 0.7514\\
Llama-3.1-70B & 0.8129 & 0.8245 & 0.4635 & 0.5410 & 0.8015 & 0.8343 & 0.7956 & 0.8042 & 0.5527 & 0.6684 & \textbf{0.8185} & \textbf{0.8612} \\
Llama-3.1-405B & 0.8485 & 0.8426 & 0.5522 & 0.6149 & 0.8615 & 0.8940 & 0.8543 & 0.8698 & 0.6421 & 0.7456 & \textbf{0.8654} & \textbf{0.9215} \\
GPT-5.5 & 0.8394 & 0.8552 & 0.5847 & 0.6420 & \textbf{0.8926} & 0.9214 & 0.8713 & 0.8859 & 0.6725 & 0.7850 & 0.8845 & \textbf{0.9412} \\
\bottomrule
\end{tabular}
}
\end{table*}

\begin{table*}[t]
\centering
\scriptsize
\caption{Supporting QA results on QuALITY. Bold marks the best method separately for Overall and Pass@5 within each backbone row.}
\label{tab:quality-results}
\setlength{\tabcolsep}{1.25pt}
\resizebox{\textwidth}{!}{%
\begin{tabular}{l|cc|cc|cc|cc|cc|cc}
\toprule
\multirow{2}{*}{\textbf{Backbone}} &
\multicolumn{2}{c|}{\textbf{Hybrid RAG}} &
\multicolumn{2}{c|}{\textbf{GraphRAG}} &
\multicolumn{2}{c|}{\textbf{LightRAG}} &
\multicolumn{2}{c|}{\textbf{HippoRAG}} &
\multicolumn{2}{c|}{\textbf{A-RAG}} &
\multicolumn{2}{c}{\textbf{NKW (Ours)}} \\
& \textbf{Overall} & \textbf{Pass@5}
& \textbf{Overall} & \textbf{Pass@5}
& \textbf{Overall} & \textbf{Pass@5}
& \textbf{Overall} & \textbf{Pass@5}
& \textbf{Overall} & \textbf{Pass@5}
& \textbf{Overall} & \textbf{Pass@5} \\
\midrule
Qwen3-8B & 0.5525 & 0.5589 & 0.5129 & 0.6132 & \textbf{0.7039} & \textbf{0.8018} & 0.5347 & 0.6211 & 0.5030 & 0.6155 & 0.6845& 0.7812\\
Qwen3-30B & 0.6215 & 0.6282 & 0.5929 & 0.6924 & 0.7535 & 0.8256 & 0.5577 & 0.5600 & 0.5886 & 0.6468 & \textbf{0.7582} & \textbf{0.8652} \\
Qwen3-235B & 0.6552 & 0.6694 & 0.6659 & 0.7662 & \textbf{0.8343} & 0.8664 & 0.6544 & 0.6619 & 0.6564 & 0.7709 & 0.8296 & \textbf{0.8989} \\
Llama-3.1-8B & 0.5319 & 0.5359 & 0.5343 & 0.6318 & \textbf{0.6425} & \textbf{0.6853} & 0.4415 & 0.4483 & 0.4851 & 0.5985 & 0.6120& 0.6639\\
Llama-3.1-70B & 0.6346 & 0.6409 & 0.6726 & 0.7749 & 0.7852 & 0.8129 & 0.6124 & 0.6231 & 0.6219 & 0.7352 & \textbf{0.7883} & \textbf{0.8541} \\
Llama-3.1-405B & 0.6421 & 0.6548 & 0.7182 & 0.8125 & \textbf{0.8216} & 0.8541 & 0.7027 & 0.7150 & 0.7111 & 0.8197 & 0.8173 & \textbf{0.8914} \\
GPT-5.5 & 0.6417 & 0.6552 & 0.7356 & 0.8411 & 0.8549 & 0.8914 & 0.7241 & 0.7384 & 0.7329 & 0.8458 & \textbf{0.8617} & \textbf{0.9231} \\
\bottomrule
\end{tabular}
}
\end{table*}

FairytaleQA and QuALITY differ from STAGE: LightRAG is strong on these shorter or more
passage-centered benchmarks, especially for one-shot QuALITY accuracy.  NKW remains competitive and
is strongest in several higher-capacity settings, with the best FairytaleQA Pass@5 for
Qwen3-30B, Qwen3-235B, Llama-3.1-70B, Llama-3.1-405B, and GPT-5.5, and the best QuALITY Pass@5 for
Qwen3-30B, Qwen3-235B, Llama-3.1-70B, Llama-3.1-405B, and GPT-5.5.  Thus,
narrative-structured evidence helps most when questions require multiple constraints, while
lightweight retrieval remains effective for local or passage-centered cases.  Appendix~\ref{sec:tool-use-appendix}
isolates this pattern by varying the allowed follow-up rounds after mandatory first-pass recall:
STAGE benefits strongly from deeper agentic interaction, whereas FairytaleQA and QuALITY show
smaller, mixed effects.  Appendix~\ref{app:qa-eval-details} gives the evaluator protocol and
agreement analysis.

\subsection{Tool-Use Analysis}
\label{sec:tool-use-analysis}

To make NKW's multi-step reasoning auditable, we group query-time tool calls by text, graph, narrative,
and document--chunk mapping channels.  Each query first runs a forced multi-channel recall step
before the agent decides whether additional tools are needed.  This forced step calls the default
graph--fact and narrative-semantic recall tools listed in Appendix~\ref{app:qa-procedure}, followed
by internal evidence assembly and source grounding; Appendix~\ref{app:evidence-binding} details the
policy.  We therefore report only the agent-selected follow-up budget: at most two rounds with up
to five tool calls per round.

\begin{figure*}[!b]
  \centering
  \includegraphics[width=0.82\linewidth]{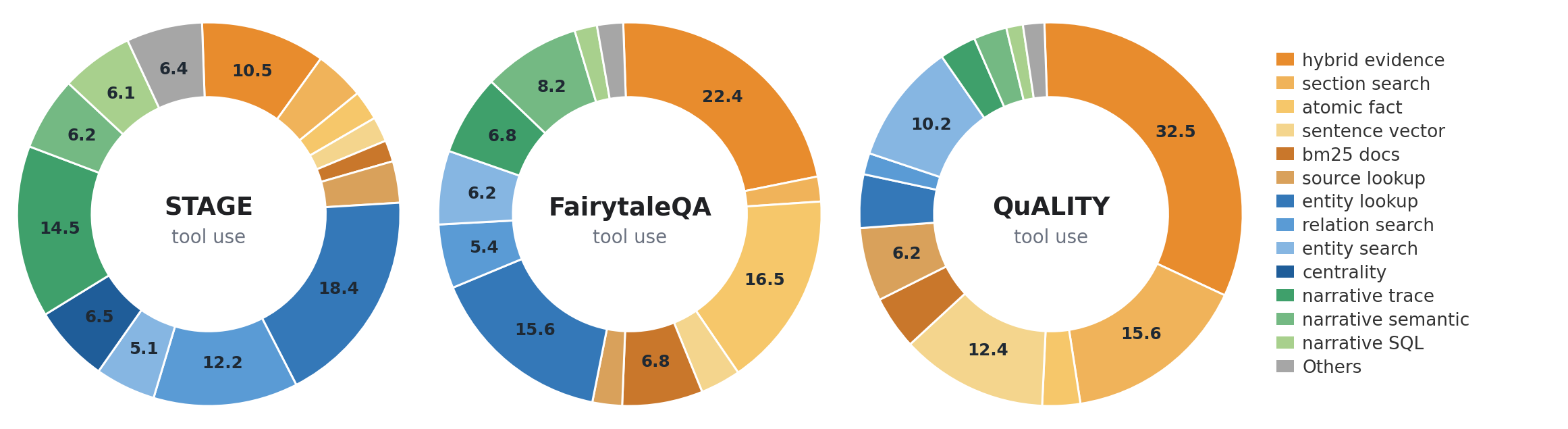}
  \caption[Fine-grained tool-use distribution across benchmarks]{\textbf{Fine-grained tool-use distribution across benchmarks.} Donut charts summarize query-time tool invocations grouped into text, graph, narrative, and administrative mapping channels. STAGE uses more graph and narrative tracing, FairytaleQA relies more on atomic facts and semantic search, and QuALITY primarily uses text extraction tools.}
  \label{fig:tool-usage}
\end{figure*}

\begin{table*}[t]
\centering
\scriptsize
\caption{Component ablations across benchmarks using Qwen3-235B. Gray subscripts indicate the absolute performance drop compared to the full system.}
\label{tab:component-ablation}
\setlength{\tabcolsep}{3.5pt}
\resizebox{\textwidth}{!}{%
\begin{tabular}{l|cc|cc|cc}
\toprule
\multirow{2}{*}{\textbf{Variant}} &
\multicolumn{2}{c|}{\textbf{STAGE}} &
\multicolumn{2}{c|}{\textbf{FairytaleQA}} &
\multicolumn{2}{c}{\textbf{QuALITY}} \\
& \textbf{Overall} & \textbf{Pass@5} 
& \textbf{Overall} & \textbf{Pass@5} 
& \textbf{Overall} & \textbf{Pass@5} \\
\midrule
Full system & 0.7012 & 0.8148 & 0.8752 & 0.9545 & 0.8296 & 0.8989 \\
\midrule
w/o attributes & 0.6715 \textcolor{gray}{$_{\downarrow 0.0297}$} & 0.7721 \textcolor{gray}{$_{\downarrow 0.0427}$} & 0.8512 \textcolor{gray}{$_{\downarrow 0.0240}$} & 0.9215 \textcolor{gray}{$_{\downarrow 0.0330}$} & 0.7816 \textcolor{gray}{$_{\downarrow 0.0480}$} & 0.8712 \textcolor{gray}{$_{\downarrow 0.0277}$} \\
w/o ep./storyline agg. & 0.6481 \textcolor{gray}{$_{\downarrow 0.0531}$} & 0.7415 \textcolor{gray}{$_{\downarrow 0.0733}$} & 0.8604 \textcolor{gray}{$_{\downarrow 0.0148}$} & 0.9312 \textcolor{gray}{$_{\downarrow 0.0233}$} & 0.7535 \textcolor{gray}{$_{\downarrow 0.0761}$} & 0.8423 \textcolor{gray}{$_{\downarrow 0.0566}$} \\
w/o graph refinement & 0.6651 \textcolor{gray}{$_{\downarrow 0.0361}$} & 0.7681 \textcolor{gray}{$_{\downarrow 0.0467}$} & 0.8488 \textcolor{gray}{$_{\downarrow 0.0264}$} & 0.9156 \textcolor{gray}{$_{\downarrow 0.0389}$} & 0.7681 \textcolor{gray}{$_{\downarrow 0.0615}$} & 0.8589 \textcolor{gray}{$_{\downarrow 0.0400}$} \\
w/o graph tools & 0.6395 \textcolor{gray}{$_{\downarrow 0.0617}$} & 0.7150 \textcolor{gray}{$_{\downarrow 0.0998}$} & 0.8445 \textcolor{gray}{$_{\downarrow 0.0307}$} & 0.8924 \textcolor{gray}{$_{\downarrow 0.0621}$} & 0.7126 \textcolor{gray}{$_{\downarrow 0.1170}$} & 0.8014 \textcolor{gray}{$_{\downarrow 0.0975}$} \\
w/o reading skills & 0.6541 \textcolor{gray}{$_{\downarrow 0.0471}$} & 0.7551 \textcolor{gray}{$_{\downarrow 0.0597}$} & 0.8556 \textcolor{gray}{$_{\downarrow 0.0196}$} & 0.9230 \textcolor{gray}{$_{\downarrow 0.0315}$} & 0.7428 \textcolor{gray}{$_{\downarrow 0.0868}$} & 0.8315 \textcolor{gray}{$_{\downarrow 0.0674}$} \\
\bottomrule
\end{tabular}
}
\end{table*}

Figure~\ref{fig:tool-usage} summarizes tool allocation across benchmarks; Appendix~\ref{app:tools-implementation}
and Tables~\ref{tab:fine-grained-tool-usage}--\ref{tab:passage-depth} give the tool taxonomy,
full distribution, autonomous follow-up depth, and follow-up-depth ablations.  The patterns match benchmark demands: STAGE
needs more graph and narrative tracing for scene order, character state, causal links, and plot
progression; FairytaleQA uses more atomic facts and semantic search; and QuALITY emphasizes
text-channel tools for option admission.  Follow-up-depth ablations further show that NKW's
agent-selected follow-up rounds bring the largest gains on STAGE.  On FairytaleQA and QuALITY, the
effect is smaller and appears mainly in the Pass@5 ceiling rather than consistent one-shot Overall
accuracy, which is expected for more passage-centered tasks.

Post-retrieval reading skills show a similar benchmark-specific pattern: STAGE triggers more quest
reasoning and character mind-state tracking, FairytaleQA relies more on literal fact localization,
and QuALITY is dominated by multiple-choice option admission.  The full distribution is reported in
Appendix~\ref{app:reading-skills}.

\subsection{Ablation Study}
\label{sec:ablation-study}

We ablate core components across STAGE, FairytaleQA, and QuALITY.  STAGE further supports
question-type analysis, linking each component to the reasoning behavior it is designed to support.

Table~\ref{tab:component-ablation} isolates five choices: entity attributes for character states,
goals, affiliations, and persistent relations; episode/storyline aggregation for long-range
progression; graph refinement for normalized entities and consolidated relations; graph-channel
tools for structured access beyond flat retrieval; and reading skills for post-retrieval answer
selection.  The \emph{w/o graph tools} variant removes only these graph-channel tools from the
available tool set; it is distinct from the zero-follow-up setting in Appendix~\ref{sec:tool-use-appendix},
which disables all autonomous follow-up rounds after the mandatory first-pass recall.

Aggregate ablations test whether each component contributes beyond one dataset.  Appendix~\ref{app:stage-qtype-ablation}
breaks STAGE down by question type in Table~\ref{tab:stage-qtype-ablation}, connecting local
components such as entity attributes to character and role-relation questions and global components
such as episode/storyline aggregation to temporal and narrative-progression reasoning.

Appendix~\ref{sec:tools-and-usage} gives method and query-time reasoning details,
Appendix~\ref{app:qa-eval-details} gives evaluation details and additional STAGE analyses,
Appendix~\ref{sec:tool-use-appendix} reports tool-use statistics, and
Appendix~\ref{sec:downstream-tasks} provides downstream application details.

\subsection{Downstream Application}
\label{sec:main-downstream-application}

\begin{figure}[t]
  \centering
  \includegraphics[width=\columnwidth]{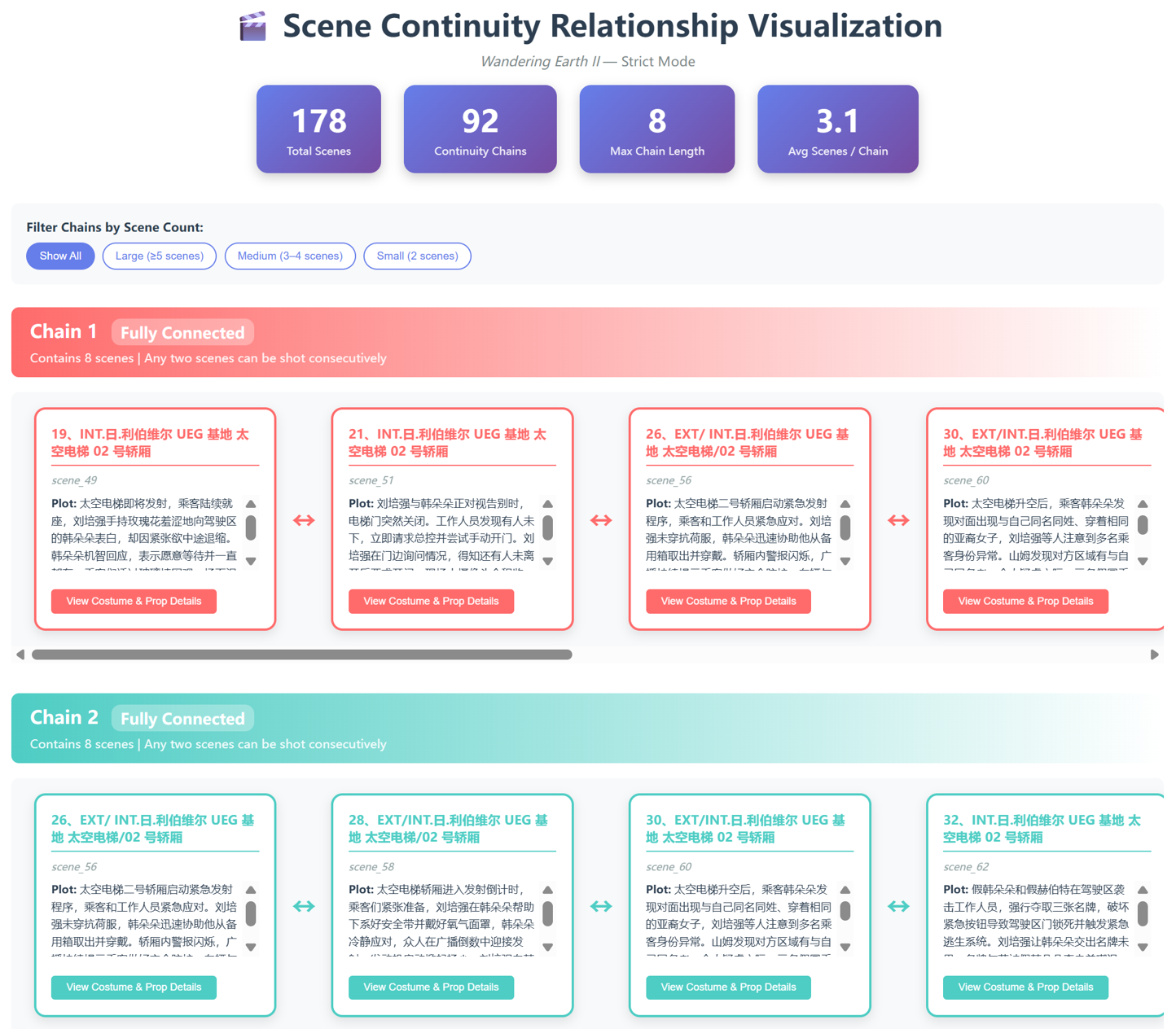}
  \caption{\textbf{Continuity chain visualization.} Chains link scenes that can plausibly share a
  production setup in an example from \textit{The Wandering Earth II}~\citep{gwo2023wanderingearth2},
  with scene summaries and source-grounded narrative cues shown for review.}
  \label{fig:continuity-visualization}
\end{figure}

The screenplay setting in STAGE evaluates not only plot comprehension, but also questions whose
evidence concerns scene structure, character presence, objects, locations, and production-relevant
state changes.  This makes it a natural testbed for whether source-grounded narrative representations
can be reused as production-facing assets rather than only as QA evidence.  We illustrate this with
production continuity checking.  Because NKW grounds entities, relations, and narrative units to
source chunks, each asset can be mapped back to the scene in which it appears; because the schema
separates characters, objects, locations, occasions, and narrative units, an agent can compare
whether two scenes are likely to share a production setup.  The resulting continuity chains can help
production teams identify compatible scenes for reusing sets, dressing, props, costumes, and makeup
baselines.  Figure~\ref{fig:continuity-visualization} shows the resulting view; the full pipeline
and prompts are in Appendix~\ref{app:continuity}, with character-state tracking in
Appendix~\ref{app:character-status}.

\section{Conclusion}
\label{sec:conclusion}

We presented \textbf{Narrative Knowledge Weaver}, a framework for source-grounded narrative knowledge
modeling and long-form QA.  NKW aligns factual, graph, entity-profile, interaction, episode, and
storyline layers so query-time reasoning can move between local evidence and higher-level narrative
progression.  It targets questions whose answers depend not on a single passage, but on how events,
character states, relations, and plot threads evolve across the story.  Across STAGE, FairytaleQA,
and QuALITY, results and ablations show that narrative-centric structure, graph-channel tools, and
reading skills improve multi-step story understanding while preserving source links.  Overall, the
findings suggest that narrative QA benefits from modeling how evidence functions inside a story,
not only where relevant passages or entity relations occur, while future work should improve
efficiency, stability, domain coverage, and source-grounded auditability.

\section*{Limitations}

NKW is designed for long-form narrative understanding, and its advantages are strongest when
questions require evidence across character states, events, temporal order, causal relations, and
story progression.  For shorter or more passage-centered questions, the additional narrative
structure can be less beneficial than lightweight retrieval, as reflected by the strong performance
of some baselines on FairytaleQA and QuALITY.  The method also incurs higher construction and
query-time cost than flat retrieval because it builds source-grounded narrative layers and performs
multi-step evidence assembly.

The quality of the constructed asset bundle depends on upstream extraction and normalization.  If
events, entity identities, or source-grounded facts are missed or incorrectly merged, later
retrieval and reasoning may inherit these errors.  Although final answers are tied back to source
evidence, the query-time agent can still select different evidence paths across runs, leading to
variance in single-sample answers.  Reading skills should also be viewed as an exploratory
component: we evaluate their aggregate contribution, but do not provide a detailed study of skill
design, selection, or interaction effects.  Our evaluation focuses on QA over existing narrative
datasets; we do not claim that the same configuration is sufficient for all narrative tasks,
domains, or languages.

\section*{Ethics Statement}

This work uses existing scientific artifacts: public narrative and long-context QA datasets,
language-model backbones, and retrieval or graph-RAG baselines.  We cite the creators of the
datasets and methods used in the experimental setup and related work.  We follow the licenses or
access terms of the original artifacts and use them for research evaluation; we do not redistribute
original benchmark texts or proprietary model weights as part of this paper.  The datasets consist
of fictional or published narrative texts and benchmark questions; we do not recruit human
participants or annotators, collect new human-subject data, or collect personally identifying
information from participants.  Some source narratives may contain fictional violence, offensive
language, stereotypes, or other sensitive content inherited from the original works.  Our use is
limited to research evaluation of narrative understanding systems.

The main risk of NKW is misuse as a tool for generating unsupported interpretations of narrative
sources or for over-trusting automatically constructed story graphs.  We mitigate this risk by
keeping graph, factual, episode, and storyline records linked to source evidence and by requiring
final answers to be grounded in original references.  The system is intended for research and
analysis settings rather than autonomous decision making about real people.  We used AI assistants
for research support, coding, and writing assistance; all paper content, experiments, and final
claims were reviewed and edited by the authors.

\bibliography{paper}

\newpage
\appendix
\numberwithin{figure}{section}
\numberwithin{table}{section}
\renewcommand{\thefigure}{\Alph{section}.\arabic{figure}}
\renewcommand{\thetable}{\Alph{section}.\arabic{table}}

\section*{Appendix Table of Contents}
\addcontentsline{toc}{section}{Appendix Table of Contents}

\begin{itemize}
  \item \textbf{Appendix~\ref{sec:tools-and-usage}: Method Details and Query-Time Reasoning}
    \begin{itemize}
        \item A.1~Narrative Aggregation Details (\S\ref{app:narrative-aggregation-details})
        \item A.2~End-to-End QA Procedure and Tool Interface (\S\ref{app:qa-procedure})
        \item A.3~Evidence Binding and Tool Policy (\S\ref{app:evidence-binding})
        \item A.4~Post-Retrieval Reading Skills (\S\ref{app:reading-skills})
        \item A.5~Multiple-Choice Option Admission (\S\ref{app:mcq-option-admission})
    \end{itemize}

  \item \textbf{Appendix~\ref{app:qa-eval-details}: Evaluation Details and Additional Results}
    \begin{itemize}
        \item B.1~Benchmark Suite Overview (\S\ref{app:benchmark-suite-details})
        \item B.2~LLM-Evaluated Correctness Prompt (\S\ref{app:llm-correctness-prompt})
        \item B.3~Evaluator Agreement (\S\ref{app:evaluator-agreement})
        \item B.4~Evaluator Stochasticity (\S\ref{app:evaluator-stochasticity})
        \item B.5~STAGE Question-Type Breakdown (\S\ref{app:stage-question-type})
        \item B.6~STAGE Question-Type Ablation (\S\ref{app:stage-qtype-ablation})
        \item B.7~STAGE Case Analysis (\S\ref{app:stage-case-analysis})
        \item B.8~Graph Asset Statistics and Computational Cost (\S\ref{app:stage-graph-assets})
        \item B.9~SABER Setting Ablation (\S\ref{app:saber-setting-ablation})
    \end{itemize}

  \item \textbf{Appendix~\ref{sec:tool-use-appendix}: Query-Time Tool-Use Analysis}
    \begin{itemize}
        \item C.1~Fine-Grained Tool-Use Distribution (\S\ref{tab:fine-grained-tool-usage})
        \item C.2~Agentic Follow-Up Depth (\S\ref{tab:tool-followup-rate})
        \item C.3~Follow-Up Depth on STAGE (\S\ref{tab:reasoning-depth})
        \item C.4~Follow-Up Depth on Passage-Centered Benchmarks (\S\ref{tab:passage-depth})
    \end{itemize}

  \item \textbf{Appendix~\ref{sec:downstream-tasks}: Downstream Applications}
    \begin{itemize}
        \item D.1~Production Continuity Checking (\S\ref{app:continuity})
        \item D.2~Character State Tracking (\S\ref{app:character-status})
    \end{itemize}
\end{itemize}

\section{Method Details and Query-Time Reasoning}
\label{sec:tools-and-usage}

This appendix collects method details that support the construction-time narrative hierarchy and
the query-time QA pipeline.  It first expands the episode/storyline aggregation procedure, then
describes evidence assembly, query-time tool interfaces, evidence binding, reading skills, and
multiple-choice option admission.

\subsection{Narrative Aggregation Details}
\label{app:narrative-aggregation-details}

\begin{figure*}[t]
  \centering
  \includegraphics[width=0.98\linewidth]{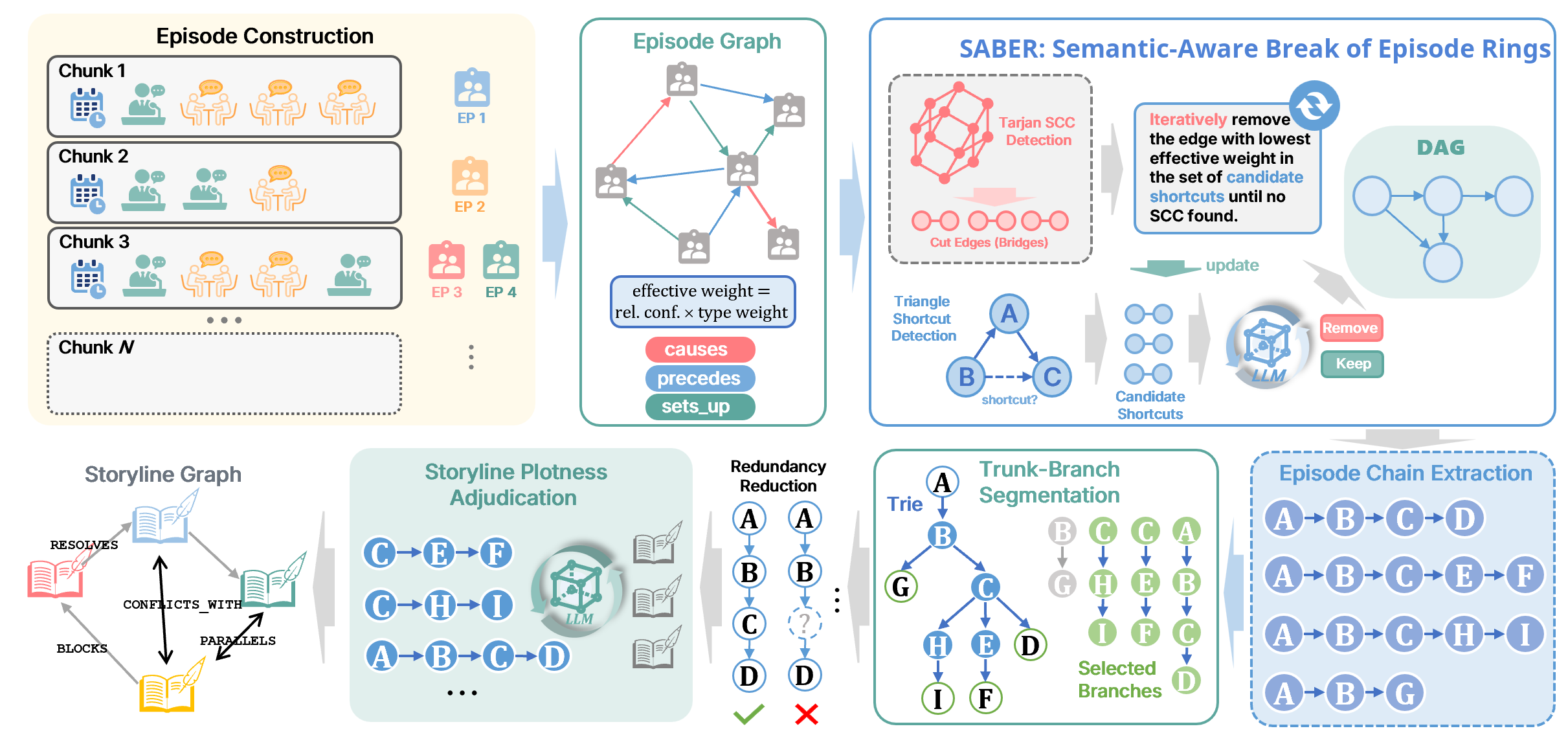}
  \caption{\textbf{Narrative aggregation from local units to storylines.}
  NKW first assembles events, interactions, and occasions into coherent episodes, extracts and
  cleans episode relations into an acyclic progression graph, derives episode chains through
  trunk--branch segmentation, and finally induces storyline nodes, support edges, and storyline
  relations.}
  \label{fig:narrative-aggregation}
\end{figure*}

Figure~\ref{fig:narrative-aggregation} expands the narrative aggregation module summarized in
Section~\ref{sec:narrative-hierarchy}.  The module starts from three kinds of base narrative units:
\emph{events}, which represent actions, discoveries, changes, conflicts, or plan outcomes;
\emph{interactions}, which represent directed exchanges among participants such as warning,
deceiving, helping, threatening, persuading, or hiding information; and \emph{occasions}, which
represent stable narrative contexts such as scenes, social situations, institutional constraints,
or recurring settings.  These units provide the local evidence from which larger narrative
structures are induced.

\paragraph{Episode assembly.}
An episode is a coherent mid-level narrative segment, not a single chunk label.  It may be local to
one scene or span multiple sections when its units jointly express the same goal, conflict, action
chain, information change, recurring occasion, or participant-centered situation.  Candidate
episode clusters are assembled using shared participants, shared goals or conflicts, causal
dependency, temporal continuity, recurring occasions, semantic coherence, source-span adjacency, and
explicit evidence links.  Each episode records its child event, interaction, and occasion units,
main participants, source spans, central goal or conflict when available, and source-grounded
evidence.  This preserves the path from an episode back to the local units and passages that
support it.

\paragraph{Episode relation graph and DAG cleaning.}
After episode assembly, NKW extracts relations among episodes.  The main directed relations are
\emph{precedes}, \emph{causes}, and \emph{elaborates}; contrastive relations are kept as auxiliary
semantic links rather than as main progression edges.  Each relation has a confidence score and a
type-dependent weight, producing an effective edge weight used during graph cleaning.  The raw
episode graph can contain cycles and redundant shortcuts.  NKW therefore converts it into a clean
episode DAG using SABER, a semantic-aware procedure for breaking cycles and pruning redundant
shortcuts in event-relation graphs.  The procedure marks triangle shortcut candidates, repeatedly
detects strongly connected components, removes the weakest edge inside each cyclic component while
preferring edges already marked as redundant shortcuts, and then re-checks remaining shortcut
candidates on the resulting DAG.  Only the remaining ambiguous shortcuts are passed to an LLM
adjudicator, which decides whether a direct edge is fully covered by an indirect path.

\begin{algorithm}[H]
\footnotesize
\caption{SABER Episode DAG Cleaning}
\label{alg:saber-dag}
\begin{algorithmic}[1]
\REQUIRE Noisy episode graph $G=(V,E)$; edge type $\tau(e)$; confidence $c(e)$; type weight $w_{\tau}$
\ENSURE Clean episode DAG $G_{\mathrm{dag}}$
\STATE Compute effective weight $s(e)=c(e)\cdot w_{\tau(e)}$ for every edge.
\STATE Mark triangle-shortcut candidates $(u,v)$ when a path $u\rightarrow x\rightarrow v$ exists.
\WHILE{$G$ contains a strongly connected component $C$ with $|C|>1$}
  \STATE Let $E_C=\{(u,v)\in E: u,v\in C\}$.
  \STATE Remove the edge in $E_C$ with minimum $s(e)$, preferring marked shortcut candidates.
  \STATE Recompute strongly connected components.
\ENDWHILE
\STATE Recompute shortcut candidates on the acyclic graph.
\FORALL{ambiguous direct edge $(u,v)$ with an indirect path $u\leadsto v$}
\STATE Ask whether $(u,v)$ is \textsc{redundant}, \textsc{necessary}, or \textsc{uncertain}.
  \IF{the decision is \textsc{redundant} and source support is preserved}
    \STATE Remove $(u,v)$.
  \ENDIF
\ENDFOR
\STATE \textbf{return} the resulting DAG $G_{\mathrm{dag}}$.
\end{algorithmic}
\end{algorithm}

The LLM is therefore used only after the graph skeleton is acyclic.  It cannot create episodes, add
new edges, reverse episode order, or remove a support link unless the direct relation is judged
redundant with respect to an already existing indirect path.

\paragraph{Chain extraction and storyline construction.}
Storyline candidates are extracted from the clean episode DAG as episode chains.  We use a
trunk--branch view rather than relying only on a longest path: source-to-sink paths are enumerated
under bounded depth and path-count constraints, inserted into a trie, and segmented into shared
trunks and meaningful branches.  This representation exposes common narrative prefixes and later
divergences, which is useful for questions about plot alternatives, delayed consequences, and
recurring arcs.  A storyline is then induced from a selected episode chain as a higher-level
narrative arc with a name, description, impact, key characters, source spans, and child episode
links.  Storyline support edges are hierarchical containment links from storylines to episodes,
whereas storyline relations describe high-level interactions among arcs, such as enabling,
blocking, resolving, conflicting with, or paralleling another storyline.

\begin{algorithm}[H]
\footnotesize
\caption{Trunk--Branch Storyline Segmentation}
\label{alg:trunk-branch}
\begin{algorithmic}[1]
\REQUIRE Clean episode DAG $G_{\mathrm{dag}}$; maximum chain length $L$; overlap threshold $\rho$; minimum new episodes $m$
\ENSURE Storyline nodes and support edges
\STATE Enumerate bounded source-to-sink episode paths in $G_{\mathrm{dag}}$.
\STATE Insert each path into a trie whose nodes are episode ids.
\STATE Extract candidate chains from shared trunks, branch continuations, and terminal paths.
\STATE Split candidates longer than $L$ at branch or low-support cut points.
\STATE Rank candidates by coverage, coherence, source support, and new episode contribution.
\STATE Initialize selected chain set $\mathcal{S}\leftarrow\emptyset$.
\FORALL{candidate chain $P$ in ranked order}
  \IF{$\max_{Q\in\mathcal{S}}\mathrm{overlap}(P,Q)\leq \rho$ and $P$ contributes at least $m$ new episodes}
    \STATE Add $P$ to $\mathcal{S}$.
  \ENDIF
\ENDFOR
\FORALL{$P\in\mathcal{S}$}
\STATE Induce a storyline with name, summary, impact, entities, source spans, and child episode ids.
  \STATE Add support edges from the storyline to its child episodes.
\ENDFOR
\STATE Predict closed-set storyline relations only between selected storylines.
\end{algorithmic}
\end{algorithm}

The complete aggregation path is therefore: source evidence to base narrative units, base units to
episodes, episodes to a clean episode DAG, and episode chains to storylines.
This hierarchy is designed to support both high-level narrative retrieval and source-grounded
answering: a query can retrieve a storyline or episode for plot-level orientation, then follow
support links back to the exact local units and passages needed for verification.

\subsection{End-to-End QA Procedure and Tool Interface}
\label{app:qa-procedure}
\label{app:tools-implementation}

Given a benchmark question and its associated document or screenplay, NKW answers the question
through a fixed evidence-assembly and finalization pipeline.  This section first summarizes the
pipeline, then details the evidence-board budget and the query-time tool interface exposed to the
reasoning agent.

\paragraph{Pipeline overview.}
\begin{enumerate}[leftmargin=*]
    \item \textbf{Query decomposition.}  The evaluation wrapper submits the question, document
    identifier, answer format, and retrieval budget to NKW.  The query is decomposed into retrieval
    intents over canonical entities, relations, atomic facts, and narrative units, so the first
    evidence pass can address both local factual grounding and higher-level narrative context.
    \item \textbf{Forced graph--fact and narrative recall.}  Before the tool agent starts, NKW runs a
    mandatory first-pass recall step over the canonical graph, atomic facts, and semantic narrative
    units.  In the query-time tool taxonomy, this step consists of:
    \begin{quote}
    \raggedright\small
    \texttt{entity\_search};
    \texttt{entity\_lookup};
    \texttt{relation\_search};
    \texttt{atomic\_fact\_search};
    \texttt{narrative\_semantic\_search}.
    \end{quote}
    This mandatory step is excluded from agent-selected tool-use statistics.
    \item \textbf{Candidate evidence assembly.}  The system enriches retrieved entities with
    attributes and query-relevant atomic facts, keeps relation candidates from both entity- and
    relation-centered retrieval, attaches available source-text evidence through provenance indices,
    and prepares source chunks for grounding.  These assembly and indexing steps are not counted as
    separate tool calls.
    \item \textbf{Tool-agent refinement.}  The tool agent is inserted after first-pass retrieval and
    before final prompt construction.  Each round scores the current packet, optionally builds a
    pruned packet, checks answerability, and stops without tool calls when the packet is already
    answerable.  Otherwise, it plans at most five tool calls for the current evidence gaps.  The
    default budget allows at most two rounds.  This fail-closed policy prevents an already sufficient
    first-pass packet from being degraded by unnecessary tool use.
    \item \textbf{Evidence-board construction.}  Returned tool evidence can add entities, relations,
    narrative items, or compact tool-text evidence, subject to channel-specific edit switches and
    added-evidence budgets.  Dynamic packing then constructs the final prompt sections for entities,
    relations, narrative items, source text chunks, optional tool-text evidence, and references.
    Optional LLM evidence refinement compresses selected entity, relation, and narrative items; it
    is an evidence compressor, not an additional retrieval tool.
    \item \textbf{Reading-skill selection and finalization.}  Reading skills are selected after
    retrieval and packing, with at most three skills injected into the final prompt.  They do not
    retrieve new evidence; they instruct the final reader how to compare the assembled packet for
    option admission, causal/quest reasoning, character mental state, theme or implication, and
    literal fact localization.  The answer model then generates the final response.
\end{enumerate}

\paragraph{Evidence-board budgeting.}
The final evidence board uses a cluster-merge-balanced allocator for entity, narrative, and relation
items.  Each candidate item is scored using query similarity, lexical overlap, source overlap,
support from selected reference chunks, token length, channel type, and relation-pair identity.
Highly similar items or items grounded in the same source evidence are merged into clusters; the
allocator then selects a small number of representatives per cluster rather than applying a flat
top-$k$ over all structured evidence.  Items already covered by selected source chunks are
downweighted to reduce repetition, relation pairs are deduplicated, and the board preserves floors
for entity and narrative evidence so that relation items cannot consume the structured-evidence
budget.  In the stable setting, the board admits up to 8 entity items, 8 narrative items, and 16
relation items, with floors of 5 for entity and narrative items and no hard floor for relations.

Entity entries are also budgeted internally.  Before an entity is shown in the final evidence board,
NKW attaches query-relevant entity attributes and at most five atomic facts by default.  Candidate
atomic facts for the entity are ranked by question similarity, then only the top facts are retained.
This keeps fine-grained source-grounded propositions available without allowing repeated local facts
to dominate the prompt.  We also implement dynamic and cluster-aware atomic-fact budgets, but the
reported experiments use the stable per-entity top-5 setting.

\paragraph{Tool interface.}
The tool-agent refinement step exposes a compact tool surface rather than a large collection of
low-level implementation calls.  The tools are organized around textual evidence, graph evidence,
narrative-structure evidence, and document--chunk mapping.  The agent can combine these channels
when a question requires local wording, entity or relation grounding, compact factual propositions,
source index resolution, and story-level progression at the same time.

\begin{table*}[t]
\centering
\small
\setlength{\tabcolsep}{4pt}
\begin{tabular}{p{2.5cm} p{4.6cm} p{6.5cm}}
\toprule
\textbf{Channel} & \textbf{Evidence units} & \textbf{Role in query-time reasoning} \\
\midrule
Text & Passages, sentences, sections, source lookups, and atomic facts & Retrieves source-grounded wording and compact factual propositions that directly support answer candidates. \\
Graph & Entities, relations, central nodes, and local graph neighborhoods & Finds salient narrative entities, attributes, relations, and local graph neighborhoods relevant to the question. \\
Narrative & Events, interactions, episodes, storylines, and structured narrative queries & Recovers temporal progression, recurring plot threads, and multi-hop narrative evidence beyond isolated chunks. \\
Other & Document titles, chunk identifiers, and document--chunk index mappings & Resolves source-index metadata, such as mapping document titles to chunk identifiers or counting chunks within a document. \\
\bottomrule
\end{tabular}
\caption{Query-time tool channels used by the reasoning agent.}
\label{tab:tools}
\end{table*}

Table~\ref{tab:public-tools} lists the tool names exposed to the query-time agent.
Lower-level retrieval, graph, and database operations are encapsulated behind this interface.

\begin{table*}[t]
\centering
\scriptsize
\setlength{\tabcolsep}{3pt}
\begin{tabular}{p{1.55cm} p{4.0cm} p{7.7cm}}
\toprule
\textbf{Channel} & \textbf{Tool name} & \textbf{Purpose} \\
\midrule
\multirow{6}{*}{Text}
& \texttt{hybrid\_evidence\_search} & Blends lexical and semantic recall for first-pass source evidence. \\
& \texttt{section\_search} & Locates or retrieves evidence from narrative sections, scenes, or chapters. \\
& \texttt{atomic\_fact\_search} & Retrieves compact source-grounded factual propositions for local states, actions, causes, functions, and rules. \\
& \texttt{vdb\_search\_sentences} & Retrieves fine-grained semantic evidence for paraphrased local facts or implications. \\
& \texttt{bm25\_search\_docs} & Retrieves source chunks with exact or rare surface terms. \\
& \texttt{source\_lookup} & Fetches original source passages after another tool returns source identifiers. \\
\midrule
\multirow{4}{*}{Graph}
& \texttt{entity\_lookup} & Retrieves attributes, relations, and source locations for a canonical entity. \\
& \texttt{entity\_search} & Finds canonical entity or concept candidates before focused lookup. \\
& \texttt{relation\_search} & Searches semantic relation facts, neighbors, entity pairs, or co-section relations. \\
& \texttt{top\_k\_by\_centrality} & Identifies globally salient entities or concepts for follow-up verification. \\
\midrule
\multirow{3}{*}{Narrative}
& \texttt{narrative\_trace\_search} & Searches event--episode--storyline traces for temporal, causal, and trajectory reasoning. \\
& \texttt{narrative\_semantic\_search} & Performs fuzzy semantic search over events, interactions, episodes, storylines, and scene notes. \\
& \texttt{narrative\_sql\_search} & Looks up structured scene notes, dialogue, action interactions, and document-linked occurrences. \\
\midrule
Other & Others & Maps between document titles and chunk identifiers, including document-level chunk listing or counting queries. \\
\bottomrule
\end{tabular}
\caption{Query-time tool names exposed to the NKW reasoning agent.  The names align with the tool legend in Figure~\ref{fig:tool-usage}; the interface is grouped into text, graph, narrative, and document--chunk mapping channels.}
\label{tab:public-tools}
\end{table*}

\subsection{Evidence Binding and Tool Policy}
\label{app:evidence-binding}

NKW separates retrieval signals from final evidence.  Tool outputs may contain graph summaries,
atomic facts, narrative-unit summaries, or candidate passages, but the final answer is grounded in
the original source references recovered from those outputs.  When several tools point to the same
source, the source receives stronger support in the final evidence packet; when a summary conflicts
with the source passage, the source passage takes precedence.  This policy is important because
graph and narrative tools are deliberately compressed views of the story, while STAGE and QuALITY
answers often turn on exact scene wording, scope, or option polarity.

The first retrieval round is a forced graph--fact and narrative-semantic recall step designed to
establish canonical entities, relations, compact factual support, and semantically relevant
narrative units before the agent chooses any follow-up tools.  The reported tool calls are the same
mandatory recall set listed in Appendix~\ref{app:qa-procedure}.  The forced round is excluded from
the agent-selected tool-use statistics; follow-up calls refine the evidence by checking exact
passages, resolving entity identity, or expanding a narrative trace.  In this design, atomic facts
and narrative traces play different roles.  Atomic facts are compact local propositions
about actions, attributes, relations, or causes that remain tied to source passages.  Narrative
traces instead organize events, episodes, storylines, and their relations, making them better suited
for temporal, causal, and trajectory questions.

We also distinguish two narrative retrieval modes.  Trace-oriented retrieval follows structured
event--episode--storyline links and is used when the question asks about order, motivation,
consequence, or a character/object trajectory.  Semantic narrative retrieval searches narrative
units by topical similarity and is more useful when the query is underspecified or when the desired
unit is a paraphrase of the question.  Keeping these modes separate reduces a common failure mode in
which a topically related scene is retrieved but the answer requires a specific earlier or later
narrative step.

\subsection{Post-Retrieval Reading Skills}
\label{app:reading-skills}

The query-time agent also uses reading skills after evidence has been retrieved and before the final
answer is generated.  These skills are not retrieval tools and do not issue additional retrieval
queries.  They are compact evidence-reading modules that tell the agent what kind of reading
operation the question requires and how retrieved evidence should be compared.

We implement these modules as literature-seeded Reading Skill Cards.  Each card is distilled from a
deterministic seed into a scoped operator with trigger signals, evidence-use rules, common failure
modes, and a compact runtime instruction.  The seeds draw on cognitive question-answering and
narrative-comprehension theories, including QUEST-style answer paths~\citep{graesser1990quest,graesser1992quest},
event-indexed situation models~\citep{zwaan1995construction,zwaan1998situation}, causal-network
accounts of story events~\citep{trabasso1985causal}, story grammar~\citep{mandler1977remembrance},
and theory-of-mind work on narrative comprehension~\citep{kim2021theory}.  We use representative
question patterns from STAGE, FairytaleQA~\citep{xu2022fairytaleqa}, and QuALITY~\citep{pang2022quality}
only to calibrate trigger signals and common failure modes.  The cards are not trained from answer
traces and do not introduce new evidence.

\begin{table}[t]
\centering
\scriptsize
\setlength{\tabcolsep}{2.5pt}
\resizebox{\columnwidth}{!}{%
\begin{tabular}{p{2.6cm} p{1.7cm} p{5.0cm}}
\toprule
\textbf{Card} & \textbf{Trigger} & \textbf{Runtime operator} \\
\midrule
Event-indexing consistency & Temporal / scene continuity &
Check time, place, protagonist, cause, and goal alignment for each evidence row; reject same-topic but different-situation support unless the source gives an explicit continuity bridge. \\
Perspective-state ledger & Character understanding &
Track the queried holder's mental state over time: separate fact from belief, anchor changes to reveal or perception events, and use only source-grounded state evidence at the asked moment. \\
QUEST legal-path reasoning & Why / how / goal questions &
Identify the queried node and frame, then answer from the closest source-grounded legal causal or goal-plan path with correct direction and actor/time binding. \\
Literal-slot grounding & Local fact questions &
Fill the exact literal slot from the smallest direct source-grounded span in the packet; keep entity--action--object--time--place bindings intact and reject broader inferred answers. \\
Story-grammar episode slots & Trigger / attempt / outcome &
Map evidence to local episode slots, identify the slot requested by the question, and answer only from that slot; do not confuse trigger, plan, attempt, outcome, or reaction. \\
\bottomrule
\end{tabular}
}
\caption{Representative Reading Skill Card operators injected into the final-answer prompt.}
\label{tab:reading-skills}
\end{table}

\begin{table}[t]
\centering
\scriptsize
\caption{Invocation frequency of post-retrieval reading skills across the benchmark suite. Percentages indicate the fraction of questions that trigger a given skill; columns can sum to $>100\%$ because up to three skills may be selected for one question. Bold marks the largest benchmark share for each skill.}
\label{tab:reading-skill-distribution}
\setlength{\tabcolsep}{2pt}
\resizebox{\columnwidth}{!}{%
\begin{tabular}{l|ccc}
\toprule
\textbf{Reading Skill} & \textbf{STAGE} & \textbf{FairytaleQA} & \textbf{QuALITY} \\
\midrule
Multiple-choice option admission & 12.5\% & 4.2\% & \textbf{96.8\%} \\
Quest and narrative reasoning    & \textbf{76.4\%} & 62.5\% & 35.2\% \\
Character mind-state tracking    & \textbf{68.2\%} & 45.3\% & 18.5\% \\
Theme and symbolic interpretation& 28.5\% & 15.6\% & \textbf{54.3\%} \\
Literal fact localization        & 21.3\% & \textbf{74.8\%} & 11.2\% \\
\midrule
\textit{Avg. Skills per Question (0--3)} & \textit{2.07} & \textit{2.02} & \textit{2.16} \\
\bottomrule
\end{tabular}
}
\end{table}

At inference time, the agent selects a small set of applicable reading skills from the question,
answer format, and retrieved evidence sketch.  The selected skills constrain interpretation but do
not replace source evidence: final answers must still be grounded in retrieved passages or their
linked narrative-graph evidence.  This separation keeps retrieval, evidence organization, and
interpretation distinct, which is important for narrative questions where the relevant evidence may
be present but easy to read with the wrong operator.

Reading skills are selected after retrieval rather than during tool use.  The selector receives the
question, answer format, and a compact sketch of the available evidence, then chooses a small number
of applicable skills from a catalog.  Required skills can be retained for special answer formats,
such as multiple-choice questions, but the selected skills remain advisory readers: they do not add
new evidence and do not override source references.  Most skills follow the same reading pattern:
identify the question frame and answer slot, map source propositions to that slot, and flag risks
such as wrong scope, unsupported actors, unsupported mechanisms, or background facts being mistaken
for the answer.

The card design keeps theory-derived claims separate from system-specific adaptations.  For example,
direct source text overriding graph summaries, limiting runtime injection to short operators, and
rejecting unsupported answer options are runtime policies for this system, not claims inherited from
the cognitive theories.  This separation makes the reading layer auditable and ablatable: disabling
reading skills removes only the final-stage evidence-reading operators while leaving retrieval and
graph construction unchanged.

\subsection{Multiple-Choice Option Admission}
\label{app:mcq-option-admission}

For multiple-choice tasks such as QuALITY, NKW uses an option-admission reader after evidence
retrieval.  The reader first converts each option into the claim it would require, then checks
whether the retrieved references directly support that claim, support it only as an inference,
leave it insufficiently supported, contradict it, or support a different relation from the one the
question asks for.  A second pass checks whether the option answers the same question grain as the
prompt: for example, whether it gives the requested cause rather than a later consequence, a broad
theme rather than a local mechanism, or a related true fact with the wrong scope.

The final answer is still produced by the answer finalizer, not by the option-admission module
alone.  The option board is used as an audit layer: options with unsupported actors, purposes,
mechanisms, scope, or polarity are discouraged even if they share surface words with the evidence.
If no option is perfectly supported, the reader marks the least unsupported choice and records the
gap.  This design is intended to reduce a common multiple-choice failure mode in long narratives,
where all options are topically plausible but only one preserves the exact source-supported
relation requested by the question.

\section{Evaluation Details and Additional Results}
\label{app:qa-eval-details}

\subsection{Benchmark Suite Overview}
\label{app:benchmark-suite-details}

The main evaluation suite combines three public narrative and long-context QA resources. STAGE
evaluates reasoning over full movie screenplays, providing one source of long, scene-structured
narrative reasoning tasks within the broader benchmark suite. QuALITY evaluates multiple-choice
reasoning over long passages where shallow retrieval is insufficient. FairytaleQA focuses on
children's stories and provides expert-authored questions over narrative elements such as
characters, settings, actions, feelings, causal relations, and outcomes.

We report STAGE with semantic correctness metrics. QuALITY is reported with
multiple-choice accuracy, and FairytaleQA is evaluated with exact or semantic answer correctness
depending on the answer format used by the official split.

\subsection{LLM-Evaluated Answer Correctness Prompt}
\label{app:llm-correctness-prompt}

To measure whether a system-generated answer is factually correct with respect to the reference
answer, we employ an LLM-based correctness evaluator.  
For each (question, system answer, reference answer) triple, the evaluator is queried independently
five times with stochastic sampling.  
Each run outputs a binary correctness label, and the final correctness score for that example is
obtained by majority voting.  
As shown in Appendix~\ref{app:evaluator-agreement}, this procedure yields high internal agreement
and provides a stable correctness signal for QA evaluation.

\paragraph{Correctness Criterion.}
The evaluator judges correctness under the following definition:

\begin{itemize}[leftmargin=1.5em]
\item \textbf{Correctness} —  
An answer is considered correct if it preserves the key factual meaning of the reference answer,
does not contradict core facts, and does not introduce hallucinated or fabricated content.
Minor variations in wording or level of detail are acceptable as long as the factual semantics
match the reference answer.
\end{itemize}

The evaluator is instructed to be conservative: if an answer is only partially correct, omits
critical information, or mixes correct facts with incorrect details, it should be labeled as
incorrect.

\begin{tcolorbox}[colback=SkyBlue!5,colframe=SkyBlue!100,halign=flush left,
title=LLM Correctness Evaluation Prompt Template, breakable=true]
\footnotesize

You are a careful factual evaluator.  
Your task is to determine whether the given answer is factually correct with respect to the
reference answer.

Judge correctness under the following rule:

\begin{itemize}[leftmargin=1.5em]
\item Label the answer as \emph{correct} if it matches the key factual meaning of the reference
answer without contradictions or hallucinations.
\item Label the answer as \emph{incorrect} if it contradicts the reference, misses essential
information, adds fabricated details, or is only partially aligned.
\end{itemize}

Return a JSON object with the structure:

\begin{verbatim}
{
  "is_correct": true | false,
  "reason": "Short judgment explanation."
}
\end{verbatim}

\vspace{4pt}
\noindent\textbf{--- Question ---}
\begin{verbatim}
{QUESTION}
\end{verbatim}

\noindent\textbf{--- System Answer ---}
\begin{verbatim}
{SYSTEM_ANSWER}
\end{verbatim}

\noindent\textbf{--- Reference Answer ---}
\begin{verbatim}
{REFERENCE_ANSWER}
\end{verbatim}

\noindent Evaluate whether the system answer is factually correct with respect to the reference
answer.  
Output only the JSON.

\end{tcolorbox}

\subsection{LLM Evaluator Agreement}
\label{app:evaluator-agreement}

To assess the stability of the LLM-based correctness evaluator
(Appendix~\ref{app:llm-correctness-prompt}), we compute its internal agreement across
five independent stochastic judgments for each (question, system answer, reference answer)
triple.  
The evaluator is queried five times, producing five binary correctness labels used for majority
voting in our evaluation.

Pairwise agreement is computed over all ${5 \choose 2}=10$ annotator pairs per example.
To quantify multi-annotator consistency, we additionally report Krippendorff's~$\alpha$ for
nominal-scale labels (Krippendorff, 2011).  
Let $n_{ic}$ denote how many times label $c\in\{0,1\}$ appears among the five judgments for
example $i$, and let $n_c$ be the total number of occurrences of label $c$ across the dataset.
The observed and expected disagreements are
\[
D_o = \frac{1}{N} 
\sum_{i=1}^N 
\frac{2\, n_{i0} n_{i1}}{5 \cdot 4},
\]
\[
D_e = 
\frac{2\, n_0 n_1}{(n_0 + n_1)^2 - (n_0^2 + n_1^2)},
\]
and Krippendorff's $\alpha$ is
\[
\alpha = 1 - \frac{D_o}{D_e}.
\]
Because $\alpha$ measures how much the annotators agree beyond what would be expected by random
labeling, it is well suited for evaluating the reliability of repeated stochastic LLM judgments.

\begin{table}[!htbp]
\centering
\caption{Internal agreement of the correctness evaluator across five independent judgments.
Pairwise agreement is averaged over all ${5 \choose 2}=10$ label pairs per example. QuALITY is
excluded because it is evaluated as multiple-choice accuracy rather than semantic answer
correctness.}
\label{tab:evaluator-agreement}
\footnotesize
\setlength{\tabcolsep}{3pt}
\resizebox{\columnwidth}{!}{%
\begin{tabular}{l|cc}
\toprule
\textbf{Dataset} & \textbf{Pairwise Agreement} & \textbf{Krippendorff's $\alpha$} \\
\midrule
STAGE & 88.5\% & 0.76 \\
FairytaleQA & 91.2\% & 0.82 \\
\bottomrule
\end{tabular}
}
\end{table}

Most disagreements arise in borderline or partially correct answers, confirming that majority
voting provides a robust and reliable estimate of semantic correctness.

\subsection{Evaluator Stochasticity}
\label{app:evaluator-stochasticity}

To validate the reliability of our LLM-as-a-judge protocol (DeepSeek-V4), we run 10 independent
evaluation passes on a random 200-question subset for each benchmark.  Table~\ref{tab:judge-stochasticity}
reports the mean overall accuracy and standard deviation across systems with different generation
paradigms.  The results show tight stability: the largest standard deviation is 0.0162, or 1.62
percentage points, across all settings.

The variance pattern is also interpretable rather than arbitrary.  NKW has the lowest variance on
both STAGE and FairytaleQA, consistent with its more structured and source-grounded answers leaving
fewer borderline cases for the judge.  GraphRAG and Hybrid RAG show slightly higher variance in some
settings, which is expected when abstractive summaries or flat-text retrieval produce partially
correct, underspecified, or over-general answers.  STAGE also shows marginally higher variance than
FairytaleQA, reflecting the greater ambiguity of screenplay-level reasoning compared with more
explicit fact-centric story questions.  Overall, these small deviations support the stability of the
evaluation protocol across distinct output formats and answer-quality regimes.

\begin{table}[!htbp]
\centering
\small
\caption{Stochasticity analysis of the LLM-as-a-judge evaluator.}
\label{tab:judge-stochasticity}
\setlength{\tabcolsep}{6pt}
\resizebox{\columnwidth}{!}{%
\begin{tabular}{l l c c}
\toprule
\textbf{Benchmark} & \textbf{System} & \textbf{Mean Acc.} & \textbf{Std ($\sigma$)} \\
\midrule
\multirow{4}{*}{STAGE}
& NKW (Full)  & 0.7015 & 0.0114 \\
& LightRAG    & 0.4412 & 0.0135 \\
& GraphRAG    & 0.4503 & 0.0162 \\
& Hybrid RAG  & 0.6251 & 0.0156 \\
\midrule
\multirow{4}{*}{FairytaleQA}
& NKW (Full)  & 0.8780 & 0.0062 \\
& LightRAG    & 0.8755 & 0.0075 \\
& GraphRAG    & 0.4485 & 0.0102 \\
& Hybrid RAG  & 0.8406 & 0.0085 \\
\bottomrule
\end{tabular}
}
\end{table}

\subsection{STAGE Question-Type Breakdown}
\label{app:stage-question-type}

Table~\ref{tab:stage-question-type} breaks down STAGE performance by its six annotated narrative
reasoning types, with all methods using the Qwen3-235B backbone.  The type distribution covers
scene grounding, character understanding, causal-motivational reasoning, temporal reasoning,
narrative progression, and role-relation continuity.

\begin{table*}[t]
\centering
\scriptsize
\caption{STAGE question-type breakdown with all methods using Qwen3-235B. Bold marks the best method separately for Overall and Pass@5 within each question type.}
\label{tab:stage-question-type}
\setlength{\tabcolsep}{1.15pt}
\resizebox{\textwidth}{!}{%
\begin{tabular}{l r|cc|cc|cc|cc|cc|cc}
\toprule
\multirow{2}{*}{\textbf{Question Type}} &
\multirow{2}{*}{\textbf{Count}} &
\multicolumn{2}{c|}{\textbf{Hybrid RAG}} &
\multicolumn{2}{c|}{\textbf{GraphRAG}} &
\multicolumn{2}{c|}{\textbf{LightRAG}} &
\multicolumn{2}{c|}{\textbf{HippoRAG}} &
\multicolumn{2}{c|}{\textbf{A-RAG}} &
\multicolumn{2}{c}{\textbf{NKW (Ours)}} \\
& &
\textbf{Overall} & \textbf{Pass@5} &
\textbf{Overall} & \textbf{Pass@5} &
\textbf{Overall} & \textbf{Pass@5} &
\textbf{Overall} & \textbf{Pass@5} &
\textbf{Overall} & \textbf{Pass@5} &
\textbf{Overall} & \textbf{Pass@5} \\
\midrule
Scene Grounding & 1207 & 0.7233 & 0.7481 & 0.2958 & 0.4582 & 0.4002 & 0.5402 & 0.4640 & 0.4814 & 0.3297 & 0.4167 & \textbf{0.7622} & \textbf{0.8575} \\
Character Understanding & 1103 & 0.7507 & 0.7679 & 0.4161 & 0.5684 & 0.6573 & 0.8205 & 0.5938 & 0.6274 & 0.5032 & 0.6102 & \textbf{0.8731} & \textbf{0.9102} \\
Causal-Motivational Reasoning & 1041 & 0.5341 & 0.5696 & 0.1988 & 0.2834 & 0.4025 & 0.5620 & 0.4131 & 0.4544 & 0.3420 & 0.4736 & \textbf{0.6436} & \textbf{0.7646} \\
Temporal Reasoning & 766 & 0.3616 & 0.4543 & 0.2598 & 0.4347 & 0.2794 & 0.4347 & 0.1632 & 0.1802 & 0.2663 & 0.4504 & \textbf{0.5339} & \textbf{0.7363} \\
Narrative Progression & 597 & 0.4456 & 0.4992 & 0.3601 & 0.6231 & 0.3333 & 0.4891 & 0.1625 & 0.1759 & 0.1893 & 0.3099 & \textbf{0.5025} & \textbf{0.6884} \\
Role-Relation Continuity & 296 & 0.7905 & 0.8446 & 0.5709 & 0.7432 & 0.5676 & 0.7872 & 0.4865 & 0.5608 & 0.6419 & 0.8007 & \textbf{0.8480} & \textbf{0.9189} \\
\midrule
\textbf{Overall} & 5010 & 0.6056 & 0.6465 & 0.3206 & 0.4790 & 0.4407 & 0.5988 & 0.4014 & 0.4301 & 0.3625 & 0.4862 & \textbf{0.7012} & \textbf{0.8148} \\
\bottomrule
\end{tabular}
}
\end{table*}

The breakdown shows that NKW improves most consistently on question types that require tracking
narrative state across non-adjacent evidence.  It is strongest on all six categories under the
Qwen3-235B setting, with especially large margins on Temporal Reasoning and Causal-Motivational
Reasoning.  The absolute scores also reveal where the benchmark remains difficult: Temporal
Reasoning and Narrative Progression remain relatively difficult, especially for systems without
explicit narrative aggregation, indicating that long-range ordering and plot-trajectory questions
remain challenging.  Hybrid RAG remains competitive on local grounding and role-relation questions, but its
Pass@5 gains are smaller than NKW's on the more global categories, suggesting that narrative
structure mainly helps when the model must assemble multiple constraints rather than retrieve a
single salient passage.

\subsection{STAGE Question-Type Ablation}
\label{app:stage-qtype-ablation}

\begin{table*}[t]
\centering
\scriptsize
\caption{Fine-grained STAGE question-type ablation using Qwen3-235B. Gray subscripts indicate the absolute performance drop compared to the full system. The degradation patterns validate the design choices: local components (e.g., attributes) disproportionately affect Character and Role-Relation tasks, while global components (e.g., storyline aggregation) are critical for Temporal and Narrative Progression reasoning.}
\label{tab:stage-qtype-ablation}
\setlength{\tabcolsep}{4pt}

\vspace{0.2cm}
\centerline{\textbf{(a) Overall Accuracy}}
\vspace{0.1cm}
\resizebox{\textwidth}{!}{%
\begin{tabular}{l|cccccc}
\toprule
\textbf{Variant} & \textbf{Character} & \textbf{Role-Relation} & \textbf{Scene} & \textbf{Causal-Motiv.} & \textbf{Temporal} & \textbf{Narrative Prog.} \\
\midrule
Full system & 0.8731 & 0.8480 & 0.7622 & 0.6436 & 0.5339 & 0.5025 \\
\midrule
w/o attributes & 0.8069 \textcolor{gray}{$_{\downarrow 0.0662}$} & 0.7905 \textcolor{gray}{$_{\downarrow 0.0575}$} & 0.7423 \textcolor{gray}{$_{\downarrow 0.0199}$} & 0.6273 \textcolor{gray}{$_{\downarrow 0.0163}$} & 0.5209 \textcolor{gray}{$_{\downarrow 0.0130}$} & 0.4891 \textcolor{gray}{$_{\downarrow 0.0134}$} \\
w/o ep./storyline agg. & 0.8568 \textcolor{gray}{$_{\downarrow 0.0163}$} & 0.8277 \textcolor{gray}{$_{\downarrow 0.0203}$} & 0.7332 \textcolor{gray}{$_{\downarrow 0.0290}$} & 0.5610 \textcolor{gray}{$_{\downarrow 0.0826}$} & 0.4073 \textcolor{gray}{$_{\downarrow 0.1266}$} & 0.4623 \textcolor{gray}{$_{\downarrow 0.0402}$} \\
w/o graph refinement & 0.8223 \textcolor{gray}{$_{\downarrow 0.0508}$} & 0.7838 \textcolor{gray}{$_{\downarrow 0.0642}$} & 0.7341 \textcolor{gray}{$_{\downarrow 0.0281}$} & 0.6100 \textcolor{gray}{$_{\downarrow 0.0336}$} & 0.5078 \textcolor{gray}{$_{\downarrow 0.0261}$} & 0.4740 \textcolor{gray}{$_{\downarrow 0.0285}$} \\
w/o graph tools & 0.8314 \textcolor{gray}{$_{\downarrow 0.0417}$} & 0.7939 \textcolor{gray}{$_{\downarrow 0.0541}$} & 0.7117 \textcolor{gray}{$_{\downarrow 0.0505}$} & 0.5668 \textcolor{gray}{$_{\downarrow 0.0768}$} & 0.4256 \textcolor{gray}{$_{\downarrow 0.1083}$} & 0.4640 \textcolor{gray}{$_{\downarrow 0.0385}$} \\
w/o reading skills & 0.8441 \textcolor{gray}{$_{\downarrow 0.0290}$} & 0.8142 \textcolor{gray}{$_{\downarrow 0.0338}$} & 0.7349 \textcolor{gray}{$_{\downarrow 0.0273}$} & 0.5783 \textcolor{gray}{$_{\downarrow 0.0653}$} & 0.4504 \textcolor{gray}{$_{\downarrow 0.0835}$} & 0.4539 \textcolor{gray}{$_{\downarrow 0.0486}$} \\
\bottomrule
\end{tabular}
}

\vspace{0.4cm}

\centerline{\textbf{(b) Pass@5}}
\vspace{0.1cm}
\resizebox{\textwidth}{!}{%
\begin{tabular}{l|cccccc}
\toprule
\textbf{Variant} & \textbf{Character} & \textbf{Role-Relation} & \textbf{Scene} & \textbf{Causal-Motiv.} & \textbf{Temporal} & \textbf{Narrative Prog.} \\
\midrule
Full system & 0.9102 & 0.9189 & 0.8575 & 0.7646 & 0.7363 & 0.6884 \\
\midrule
w/o attributes & 0.8305 \textcolor{gray}{$_{\downarrow 0.0797}$} & 0.8378 \textcolor{gray}{$_{\downarrow 0.0811}$} & 0.8235 \textcolor{gray}{$_{\downarrow 0.0340}$} & 0.7349 \textcolor{gray}{$_{\downarrow 0.0297}$} & 0.7141 \textcolor{gray}{$_{\downarrow 0.0222}$} & 0.6667 \textcolor{gray}{$_{\downarrow 0.0217}$} \\
w/o ep./storyline agg. & 0.8867 \textcolor{gray}{$_{\downarrow 0.0235}$} & 0.8818 \textcolor{gray}{$_{\downarrow 0.0371}$} & 0.8169 \textcolor{gray}{$_{\downarrow 0.0406}$} & 0.6609 \textcolor{gray}{$_{\downarrow 0.1037}$} & 0.6123 \textcolor{gray}{$_{\downarrow 0.1240}$} & 0.5578 \textcolor{gray}{$_{\downarrow 0.1306}$} \\
w/o graph refinement & 0.8522 \textcolor{gray}{$_{\downarrow 0.0580}$} & 0.8480 \textcolor{gray}{$_{\downarrow 0.0709}$} & 0.8136 \textcolor{gray}{$_{\downarrow 0.0439}$} & 0.7205 \textcolor{gray}{$_{\downarrow 0.0441}$} & 0.6997 \textcolor{gray}{$_{\downarrow 0.0366}$} & 0.6516 \textcolor{gray}{$_{\downarrow 0.0368}$} \\
w/o graph tools & 0.8404 \textcolor{gray}{$_{\downarrow 0.0698}$} & 0.8345 \textcolor{gray}{$_{\downarrow 0.0844}$} & 0.7672 \textcolor{gray}{$_{\downarrow 0.0903}$} & 0.6503 \textcolor{gray}{$_{\downarrow 0.1143}$} & 0.6110 \textcolor{gray}{$_{\downarrow 0.1253}$} & 0.5645 \textcolor{gray}{$_{\downarrow 0.1239}$} \\
w/o reading skills & 0.8722 \textcolor{gray}{$_{\downarrow 0.0380}$} & 0.8750 \textcolor{gray}{$_{\downarrow 0.0439}$} & 0.8161 \textcolor{gray}{$_{\downarrow 0.0414}$} & 0.6792 \textcolor{gray}{$_{\downarrow 0.0854}$} & 0.6723 \textcolor{gray}{$_{\downarrow 0.0640}$} & 0.5946 \textcolor{gray}{$_{\downarrow 0.0938}$} \\
\bottomrule
\end{tabular}
}
\end{table*}

The ablation results connect the aggregate component drops in Table~\ref{tab:component-ablation} to
the question types each component is designed to support.  Removing attributes mainly hurts
Character Understanding and Role-Relation Continuity, which depend on stable entity attributes,
states, affiliations, and recurring relations.  In contrast, removing episode/storyline aggregation
has its largest effect on Temporal Reasoning, Causal-Motivational Reasoning, and Narrative
Progression, showing that event-level evidence alone is insufficient for questions about ordering,
motivation, and plot trajectory.  Removing graph-channel tools causes broad degradation, especially
in Pass@5, which indicates that the query-time agent benefits from structured graph access rather
than only from the constructed graph existing in the background.  This ablation keeps the rest of the
tool-using policy available and is therefore separate from the follow-up-depth intervention in
Appendix~\ref{sec:tool-use-appendix}.  Reading skills have a smaller effect on
local categories but larger drops on causal, temporal, and progression-oriented questions, where the
retrieved evidence must be interpreted with the right temporal, causal, or discourse operator.

\subsection{STAGE Case Analysis}
\label{app:stage-case-analysis}

To complement the aggregate STAGE results, we inspect representative Qwen3-235B cases from
\textit{Birthday Girl}.  This film is useful because many questions require separating adjacent
domestic scenes, identifying the initiating event in a chain, or tracking a relationship shift
across non-contiguous scenes.  Table~\ref{tab:stage-case-analysis} summarizes cases where NKW is
correct in Pass@5 while several baselines produce majority-incorrect answers.

\begin{table*}[t]
\centering
\scriptsize
\caption{Representative STAGE cases from \textit{Birthday Girl} under Qwen3-235B.  The table
summarizes majority-incorrect baseline behavior and contrasts it with NKW's answers.}
\label{tab:stage-case-analysis}
\setlength{\tabcolsep}{3pt}
\resizebox{\textwidth}{!}{%
\begin{tabular}{p{0.11\textwidth}|p{0.13\textwidth}|p{0.18\textwidth}|p{0.31\textwidth}|p{0.24\textwidth}}
\toprule
\textbf{Case} & \textbf{Type} & \textbf{Reference} & \textbf{Baseline failure pattern} & \textbf{NKW behavior} \\
\midrule
Q2: bathroom encounter &
Character Understanding &
Nadia locks herself in with John. &
LightRAG, HippoRAG, GraphRAG, Hybrid RAG, and A-RAG are majority-incorrect.  They either report that
no decision is evidenced, retrieve a different bathroom/landing moment, or hallucinate an emotional
reaction such as Nadia sobbing. &
NKW answers correctly in 4/5 runs by grounding the local action: Nadia closes the door and remains
inside with John. \\
\midrule
Q10: tense meal &
Causal-Motivational Reasoning &
Prolonged eye contact during the meal creates the tension that leads to the ring gesture. &
HippoRAG and GraphRAG are majority-incorrect, while A-RAG over-focuses on the later ring placement
itself.  Incorrect answers often replace the silent tension with fabricated dialogue or with the
consequence rather than the preceding cause. &
NKW answers correctly in 5/5 runs by staying within the kitchen scene and connecting the ring
gesture to the charged interaction immediately before it. \\
\midrule
Q18: intimate trajectory &
Narrative Progression &
Nadia initiates physical intimacy after John's rejection. &
LightRAG, HippoRAG, GraphRAG, and Hybrid RAG are majority-incorrect.  Typical errors shift the pivot to
a violent fight, Alexei's suite, or another broad plot event, losing the local agentive action that
changes the trajectory. &
NKW answers correctly in 4/5 runs, using the narrative causal trace to identify Nadia's initiating
action as the pivot toward the intimate encounter and its aftermath. \\
\midrule
Q28: start of interaction &
Temporal Reasoning &
John greets arriving passenger Nadia at the airport before the motorway drive. &
HippoRAG, Hybrid RAG, and A-RAG are majority-incorrect.  Their answers drift to later travel events,
such as a lift exchange, a car ignition, or leaving a police station, which are temporally adjacent
but not the start of the interaction. &
NKW is correct in Pass@5 by retrieving the airport interaction and distinguishing the initiating
greeting from later movement toward the car. \\
\bottomrule
\end{tabular}
}
\end{table*}

The cases expose three recurring advantages of the narrative-centric representation.  First,
scene-level provenance reduces local collisions: in Q2, the relevant decision is a short physical
action inside the landing/bathroom sequence, and methods relying on less structured retrieval often
substitute nearby but wrong moments.  Second, event and causal traces help distinguish a cause from
its consequence: in Q10, the answer is not merely that the ring is placed, but the preceding charged
meal interaction that motivates it.  Third, storyline-level structure helps temporal and progression
questions choose the right abstraction level.  In Q18 and Q28, several baselines retrieve events that
are narratively related but occur at the wrong point in the chain; NKW more often anchors the answer
to the initiating event required by the question.

We also inspect failure cases to clarify the current boundary of NKW's reasoning behavior.
Table~\ref{tab:nkw-failure-cases} reports non-Scene-Grounding examples from diagnostic STAGE runs
over \textit{Birthday Girl} and \textit{Chasing Amy}.  These cases show that source-grounded
narrative assets reduce many retrieval failures, but do not eliminate errors in selecting the exact
causal trigger, temporal anchor, or speaker-level motivation when several semantically related
events compete.

\begin{table*}[t]
\centering
\scriptsize
\caption{Representative NKW failure cases from STAGE diagnostic runs.  We exclude Scene Grounding
questions to focus on reasoning failures rather than direct scene lookup.}
\label{tab:nkw-failure-cases}
\setlength{\tabcolsep}{3pt}
\resizebox{\textwidth}{!}{%
\begin{tabular}{p{0.11\textwidth}|p{0.14\textwidth}|p{0.23\textwidth}|p{0.29\textwidth}|p{0.19\textwidth}}
\toprule
\textbf{Case} & \textbf{Type} & \textbf{Reference} & \textbf{NKW failure pattern} & \textbf{Diagnostic implication} \\
\midrule
\textit{Birthday Girl} Q16 &
Narrative Progression &
The airport meeting initiates the shift in John and Nadia's relationship. &
NKW repeatedly selects later intimate or domestic moments, such as the bedroom encounter or ring
gesture, treating salient consequences as the initiating event. &
Progression questions can still confuse the start of a trajectory with later high-salience events
inside that trajectory. \\
\midrule
\textit{Chasing Amy} Q3 &
Character Understanding &
Holden catches up to Alyssa in order to confront her about the rumors. &
NKW answers from the broader romantic arc, saying that Holden seeks love or a relationship, rather
than the local conversational objective in the parking-lot scene. &
Character-intention questions remain vulnerable to global relationship summaries when the question
asks for a local action goal. \\
\midrule
\textit{Chasing Amy} Q16 &
Narrative Progression &
Banky's mockery of Holden's fixation on Alyssa escalates the tension between Holden and Banky. &
NKW jumps to a later dramatic confrontation involving Holden's threesome proposal and kiss, which is
narratively related but not the earlier catalyst requested by the question. &
Storyline retrieval can over-prefer dramatic downstream events unless the requested turning point is
anchored to the correct causal position. \\
\midrule
\textit{Chasing Amy} Q33 &
Causal-Motivational Reasoning &
Jay's advice that Alyssa choosing Holden now matters more than her past motivates Holden to leave
the diner. &
NKW often attributes the decision to Silent Bob's ``chasing Amy'' reflection or another nearby
emotional cue, misidentifying both the speaker and the immediate trigger. &
Dialogue-level causal attribution remains difficult when adjacent advice scenes share similar
themes but differ in speaker, timing, and consequence. \\
\bottomrule
\end{tabular}
}
\end{table*}

\subsection{STAGE Graph Asset Statistics}
\label{app:stage-graph-assets}

Table~\ref{tab:stage-graph-assets} reports the graph-asset footprint for the 40-screenplay STAGE
setting used in the Qwen3-235B experiments.  The numbers should be read as representation-scale
statistics rather than as a direct quality metric, because each system defines nodes, edges, and
auxiliary graph assets differently.  Table~\ref{tab:nkw-stage-assets} further breaks down the NKW
asset bundle into its entity-level and narrative-level components.  Table~\ref{tab:computational-cost}
reports the corresponding construction and query-time wall-clock costs, measured under the same
16-worker parallel setup.  Construction time is reported in minutes per screenplay, whereas
QA + Retrieval is reported in seconds per query.  SABER-specific ablations are reported separately in Appendix~\ref{app:saber-setting-ablation}.

\begin{table}[t]
\centering
\small
\caption{STAGE graph-asset scale across graph-based systems. Statistics are per screenplay unless
otherwise stated. Because graph schemas differ across systems, the table is intended as a footprint
comparison rather than a direct measure of graph quality.}
\label{tab:stage-graph-assets}
\setlength{\tabcolsep}{3.5pt}
\resizebox{\columnwidth}{!}{%
\begin{tabular}{llrrrr}
\toprule
\textbf{System} & \textbf{Asset} & \textbf{Mean} & \textbf{Median} & \textbf{Min} & \textbf{Max} \\
\midrule
\multirow{2}{*}{HippoRAG} 
& Nodes & 1,252.0 & 1,236.0 & 244 & 2,886 \\
& Edges & 5,552.1 & 5,145.5 & 741 & 14,048 \\
\midrule
\multirow{3}{*}{GraphRAG}
& Entity nodes & 397.8 & 276.5 & 129 & 2,078 \\
& Relationships & 946.7 & 640.0 & 266 & 5,873 \\
& Community reports & 87.9 & 57.0 & 25 & 551 \\
\midrule
\multirow{2}{*}{LightRAG}
& Nodes & 585.5 & 574.0 & 229 & 1,020 \\
& Edges & 752.2 & 710.5 & 284 & 1,334 \\
\midrule
\multirow{2}{*}{NKW (Ours)}
& KG nodes & 624.5 & 595.0 & 245 & 1,150 \\
& KG edges & 812.2 & 780.5 & 310 & 1,480 \\
\bottomrule
\end{tabular}
}
\end{table}

\begin{table}[t]
\centering
\scriptsize
\caption{Detailed NKW asset statistics on STAGE. NKW stores both a source-grounded entity graph
and higher-level narrative assets, including episodes, storylines, and their support links.}
\label{tab:nkw-stage-assets}
\setlength{\tabcolsep}{3pt}
\resizebox{\columnwidth}{!}{%
\begin{tabular}{lrrrr}
\toprule
\textbf{Graph asset} & \textbf{Avg. / screenplay} & \textbf{Median} & \textbf{Min} & \textbf{Max} \\
\midrule
KG nodes & 624.50 & 595.0 & 245 & 1,150 \\
KG edges & 812.20 & 780.5 & 310 & 1,480 \\
Refined relations & 835.40 & 795.5 & 315 & 1,520 \\
Doc-entity edges & 715.30 & 680.5 & 270 & 1,300 \\
Documents & 162.70 & 143.5 & 36 & 384 \\
Episodes & 78.50 & 75.0 & 30 & 140 \\
Episode relations & 118.20 & 112.5 & 45 & 210 \\
Episode DAG relations & 80.10 & 76.5 & 30 & 145 \\
Storylines & 8.35 & 8.0 & 3 & 15 \\
Storyline relations & 6.05 & 5.5 & 1 & 12 \\
Storyline support edges & 35.80 & 34.0 & 12 & 65 \\
\bottomrule
\end{tabular}
}
\end{table}

\begin{table}[t]
\centering
\scriptsize
\caption{Computational cost for graph construction and query-time QA on STAGE.}
\label{tab:computational-cost}
\setlength{\tabcolsep}{3.5pt}
\resizebox{\columnwidth}{!}{%
\begin{tabular}{lllrrrr}
\toprule
\textbf{System} & \textbf{Phase} & \textbf{Unit} & \textbf{Mean} & \textbf{Median} & \textbf{Min} & \textbf{Max} \\
\midrule
\multirow{2}{*}{HippoRAG}
& Construction & min. & 2.32 & 2.24 & 1.65 & 3.88 \\
& QA + Retrieval & sec. & 2.79 & 2.78 & 2.26 & 3.45 \\
\midrule
\multirow{2}{*}{GraphRAG}
& Construction & min. & 16.33 & 13.95 & 5.57 & 38.00 \\
& QA + Retrieval & sec. & 485.50 & 446.20 & 305.15 & 815.30 \\
\midrule
\multirow{2}{*}{LightRAG}
& Construction & min. & 7.46 & 6.66 & 2.96 & 16.84 \\
& QA + Retrieval & sec. & 9.15 & 8.80 & 5.92 & 18.75 \\
\midrule
\multirow{2}{*}{\textbf{NKW (Ours)}}
& Construction & min. & 10.12 & 8.95 & 3.85 & 22.50 \\
& QA + Retrieval & sec. & 36.45 & 34.20 & 18.50 & 82.50 \\
\bottomrule
\end{tabular}
}
\end{table}

The comparison clarifies the representational difference behind the QA and ablation results.
HippoRAG constructs the densest graph, especially in edge count.  GraphRAG uses fewer entity nodes
than NKW but maintains many relationships and community reports, while LightRAG keeps a smaller
entity-relation footprint.  NKW is not simply a larger flat graph: it maintains a sizeable entity
graph together with document links, refined relations, episodes, storyline nodes, and cross-level
support edges.  This multi-level asset bundle explains
why the STAGE ablations in Table~\ref{tab:stage-qtype-ablation} show different failure modes for
local entity attributes, graph-channel tools, and episode/storyline aggregation.  The intended benefit is
not graph size alone, but separating local grounding from temporal, causal, and storyline-level
evidence so the query-time agent can retrieve the right layer for each question type.  The cost
profile in Table~\ref{tab:computational-cost} shows the corresponding trade-off: NKW adds moderate
construction overhead over LightRAG because it builds narrative hierarchy, and it incurs higher
query-time latency because of agentic tool calls.  It nevertheless remains far less expensive than
GraphRAG's community-summary query pipeline.

\subsection{SABER Setting Ablation}
\label{app:saber-setting-ablation}
We isolate the effect of SABER within the narrative aggregation pipeline.  This ablation keeps
episode assembly, query tools, reading skills, and the QA backbone (Qwen3-235B) fixed, varying only
how raw episode relations are converted into the progression graph prior to chain extraction.
\textbf{Raw} uses the uncleaned episode relation graph with bounded simple-path traversal to avoid
infinite loops.  \textbf{Heuristic DAG} removes the lowest-scoring edge inside each strongly
connected component (SCC) until the graph is acyclic, but performs no semantic redundancy checks.
\textbf{SABER} applies the full two-stage procedure: it first guarantees an acyclic skeleton, then
uses LLM-late semantic pruning to remove direct edges that are already logically covered by indirect
paths.

In the 151-screenplay STAGE construction pool, only 31 screenplays contained SCCs that strictly
required cycle breaking.  Tables~\ref{tab:saber-setting-ablation} and~\ref{tab:saber-qtype-ablation}
report this 31-screenplay SCC subset, where the difference between raw traversal, heuristic cycle
breaking, and SABER is directly observable.  Redundant semantic shortcuts, however, were pervasive
across the broader corpus.  SABER therefore applies LLM-adjudicated shortcut pruning regardless of
whether the initial graph contains cycles, making it a general denoising step for storyline
extraction.  This comparison tests whether semantic-aware cycle breaking stabilizes the
induced storyline asset bundle and whether cleaner trajectories translate into better QA on
ordering-, causality-, and progression-sensitive questions.

\begin{table*}[t]
\centering
\scriptsize
\caption{SABER setting ablation on the 31 STAGE screenplays whose raw episode graphs contain SCCs.
All settings reuse the same extracted episodes and raw episode relations; only the episode-relation
cleanup strategy is changed before chain extraction and storyline construction. Asset statistics are
averaged per screenplay, and QA metrics use the Qwen3-235B backbone.}
\label{tab:saber-setting-ablation}
\setlength{\tabcolsep}{3.5pt}
\resizebox{\textwidth}{!}{%
\begin{tabular}{lrrrrrrrrr}
\toprule
\textbf{Setting} &
\textbf{Prog. Edges} &
\textbf{Remain. SCC} &
\textbf{Cycle Edges Removed} &
\textbf{Shortcut Edges Removed} &
\textbf{Storylines} &
\textbf{Avg. SL Len.} &
\textbf{SL Relations} &
\textbf{Overall} &
\textbf{Pass@5} \\
\midrule
Raw & 363.7 & 0.32 & 0.0 & 0.0 & 12.4 & 6.2 & 14.8 & 0.6712 & 0.7725 \\
Heuristic DAG & 355.2 & 0.00 & 8.5 & 0.0 & 10.1 & 7.1 & 9.5 & 0.6854 & 0.7942 \\
SABER & 246.7 & 0.00 & 8.5 & 108.5 & 8.35 & 9.4 & 6.05 & \textbf{0.7012} & \textbf{0.8147} \\
\bottomrule
\end{tabular}
}
\end{table*}

\begin{table}[t]
\centering
\scriptsize
\caption{SABER ablation on the SCC subset for STAGE question types most sensitive to episode
ordering and storyline structure. Pruning semantic shortcuts improves temporal progression and
causal-chain tracking.}
\label{tab:saber-qtype-ablation}
\resizebox{\columnwidth}{!}{%
\begin{tabular}{l|cc|cc|cc}
\toprule
\textbf{Setting} &
\multicolumn{2}{c|}{\textbf{Causal-Motiv.}} &
\multicolumn{2}{c|}{\textbf{Temporal}} &
\multicolumn{2}{c}{\textbf{Narr. Prog.}} \\
& \textbf{Overall} & \textbf{Pass@5}
& \textbf{Overall} & \textbf{Pass@5}
& \textbf{Overall} & \textbf{Pass@5} \\
\midrule
Raw & 0.6105 & 0.7120 & 0.4725 & 0.6650 & 0.4450 & 0.6120 \\
Heuristic DAG & 0.6258 & 0.7385 & 0.4985 & 0.7015 & 0.4685 & 0.6455 \\
SABER & \textbf{0.6430} & \textbf{0.7646} & \textbf{0.5298} & \textbf{0.7370} & \textbf{0.4994} & \textbf{0.6864} \\
\bottomrule
\end{tabular}
}
\end{table}

The ablation separates two effects.  The heuristic DAG removes the small number of cycle-inducing
edges and improves over the raw graph, but it still leaves many semantically redundant shortcuts.
SABER removes those shortcuts after the graph skeleton is acyclic, yielding fewer progression edges,
fewer storyline relations, and longer average storyline chains.  The QA gains are concentrated on
causal, temporal, and narrative-progression questions, which are precisely the categories that rely
on clean episode order and non-spurious long-range trajectories.

\section{Query-Time Tool-Use Analysis}
\label{sec:tool-use-appendix}

This appendix reports the query-time tool-use statistics summarized in Section~\ref{sec:tool-use-analysis}
and ablates the same control variable used at inference time: the maximum number of autonomous
follow-up rounds after the mandatory first-pass evidence recall.  All percentages, follow-up-depth
statistics, and follow-up-depth ablations exclude the mandatory recall itself; they measure only
tools selected by the reasoning agent after its answerability check.

\begin{table}[t]
\centering
\scriptsize
\caption{Fine-grained tool-use distribution across the benchmark suite.  Percentages are normalized within each benchmark. Administrative calls, such as resolving document titles from chunk identifiers or listing chunks for a document, are aggregated into the ``Others'' row.}
\label{tab:fine-grained-tool-usage}
\setlength{\tabcolsep}{3pt}
\begin{tabular}{l r r r}
\toprule
\textbf{Tool} & \textbf{STAGE} & \textbf{FairytaleQA} & \textbf{QuALITY} \\
\midrule
\multicolumn{4}{l}{\textbf{Text}} \\
\texttt{hybrid\_evidence\_search} & 10.5\% & 22.4\% & 32.5\% \\
\texttt{section\_search}          & 4.2\%  & 2.1\%  & 15.6\% \\
\texttt{atomic\_fact\_search}     & 2.5\%  & 16.5\% & 3.2\%  \\
\texttt{vdb\_search\_sentences}   & 2.1\%  & 3.4\%  & 12.4\% \\
\texttt{bm25\_search\_docs}       & 1.8\%  & 6.8\%  & 4.5\%  \\
\texttt{source\_lookup}           & 3.5\%  & 2.5\%  & 6.2\%  \\
\midrule
\multicolumn{4}{l}{\textbf{Graph}} \\
\texttt{entity\_lookup}           & 18.4\% & 15.6\% & 4.5\%  \\
\texttt{relation\_search}         & 12.2\% & 5.4\%  & 1.8\%  \\
\texttt{entity\_search}           & 5.1\%  & 6.2\%  & 10.2\% \\
\texttt{top\_k\_by\_centrality}   & 6.5\%  & --     & --     \\
\midrule
\multicolumn{4}{l}{\textbf{Narrative}} \\
\texttt{narrative\_trace\_search} & 14.5\% & 6.8\%  & 3.1\%  \\
\texttt{narrative\_semantic\_search} & 6.2\% & 8.2\%  & 2.8\%  \\
\texttt{narrative\_sql\_search}   & 6.1\%  & 1.9\%  & 1.4\%  \\
\midrule
\multicolumn{4}{l}{\textbf{Misc.}} \\
Others              & 6.4\%  & 2.2\%  & 1.8\%  \\
\bottomrule
\end{tabular}
\end{table}

\paragraph{Observed follow-up depth.}
Beyond tool-channel distribution, we measure how often the agent actually chooses additional tools
after the first-pass recall.  Table~\ref{tab:tool-followup-rate} reports this observed depth with the
Qwen3-235B backbone.  Passage-centered QuALITY rarely triggers follow-up, and FairytaleQA does so in
a minority of cases.  STAGE is substantially more interactive: 91.5\% of questions require at least
one agent-selected follow-up, with 1.35 follow-up rounds and 5.8 tool calls on average.  This pattern
matches STAGE's emphasis on scene order, character state, causal motivation, and storyline
progression, where evidence often has to be assembled across text, graph, and narrative layers.

\begin{table}[t]
\centering
\scriptsize
\caption{Agentic tool-use depth across benchmarks using Qwen3-235B. The autonomous follow-up rate
is the percentage of questions for which the reasoning agent selected additional tools after the
forced first-pass evidence recall.}
\label{tab:tool-followup-rate}
\setlength{\tabcolsep}{6pt}
\resizebox{\columnwidth}{!}{%
\begin{tabular}{lrrr}
\toprule
\textbf{Benchmark} &
\textbf{Follow-up Rate} &
\textbf{Avg. Rounds} &
\textbf{Avg. Tool Calls} \\
\midrule
STAGE       & \textbf{91.5\%} & \textbf{1.35} & \textbf{5.8} \\
FairytaleQA & 24.2\% & 0.35 & 2.4 \\
QuALITY     & 8.4\%  & 0.12 & 1.5 \\
\bottomrule
\end{tabular}
}
\end{table}

\paragraph{Follow-up-depth ablation on STAGE.}
Table~\ref{tab:reasoning-depth} changes only the maximum allowed follow-up rounds on STAGE; the
Qwen3-235B backbone, mandatory first-pass recall, evidence-packing policy, and final answer
generation are fixed.  The 0-round setting therefore tests whether first-pass evidence alone is
sufficient, not whether retrieval is disabled.  Allowing one round recovers much of the loss, and the
full 2-round setting gives the best result, especially on temporal and narrative-progression
questions.

\begin{table}[t]
\centering
\scriptsize
\caption{Follow-up-depth ablation on STAGE using Qwen3-235B.  Max Follow-up specifies how many
additional tool-invocation cycles the agent may perform after the mandatory first-pass recall.}
\label{tab:reasoning-depth}
\setlength{\tabcolsep}{3.3pt}
\resizebox{\columnwidth}{!}{%
\begin{tabular}{l|cc|cc|cc}
\toprule
\multirow{2}{*}{\textbf{Max Follow-up}} &
\multicolumn{2}{c|}{\textbf{All Types}} &
\multicolumn{2}{c|}{\textbf{Temporal}} &
\multicolumn{2}{c}{\textbf{Narr. Prog.}} \\
& \textbf{Overall} & \textbf{Pass@5}
& \textbf{Overall} & \textbf{Pass@5}
& \textbf{Overall} & \textbf{Pass@5} \\
\midrule
0 rounds & 0.6126 & 0.6840 & 0.4021 & 0.5418 & 0.3853 & 0.5209 \\
1 round & 0.6780 & 0.7754 & 0.4948 & 0.6841 & 0.4523 & 0.6181 \\
2 rounds & \textbf{0.7012} & \textbf{0.8148} & \textbf{0.5339} & \textbf{0.7363} & \textbf{0.5025} & \textbf{0.6884} \\
\bottomrule
\end{tabular}
}
\end{table}

\paragraph{Follow-up-depth ablation on passage-centered benchmarks.}
Table~\ref{tab:passage-depth} applies the same 0-round versus 2-round intervention to FairytaleQA
and QuALITY.  The effect is much smaller than on STAGE.  FairytaleQA is nearly unchanged in Overall
but improves under Pass@5 with two rounds, while QuALITY slightly favors 0 rounds in one-shot Overall
accuracy and two rounds under Pass@5.  We therefore keep the 2-round budget as the default for a
single system configuration, while noting that adaptive early stopping is attractive for shallow
passage-centered queries.

\begin{table}[t]
\centering
\scriptsize
\caption{Follow-up-depth ablation on passage-centered benchmarks using Qwen3-235B.  Max Follow-up
specifies how many additional tool-invocation cycles the agent may perform after the mandatory
first-pass recall.  Bold marks the better score within each column.}
\label{tab:passage-depth}
\setlength{\tabcolsep}{4.5pt}
\resizebox{\columnwidth}{!}{%
\begin{tabular}{l|cc|cc}
\toprule
\multirow{2}{*}{\textbf{Max Follow-up}} &
\multicolumn{2}{c|}{\textbf{FairytaleQA}} &
\multicolumn{2}{c}{\textbf{QuALITY}} \\
& \textbf{Overall} & \textbf{Pass@5}
& \textbf{Overall} & \textbf{Pass@5} \\
\midrule
0 rounds & \textbf{0.8773} & 0.9312 & \textbf{0.8351} & 0.8617 \\
2 rounds & 0.8752 & \textbf{0.9545} & 0.8296 & \textbf{0.8989} \\
\bottomrule
\end{tabular}
}
\end{table}

\section{Downstream Applications}\label{sec:downstream-tasks}
\subsection{Production Continuity Checking}\label{app:continuity}

This section describes a downstream application enabled by source-grounded scene--entity structure:
\emph{production continuity checking}.  The goal is to determine whether two scenes can plausibly
share a production setup.  We do not assume a separate production-design extractor.  Instead, the
agent uses the same assets described in the main paper: source chunks, canonical entities, linked
locations and objects, occasions, events, interactions, atomic facts, and scene titles.  Visible
production cues are considered only when they are supported by these source-grounded assets.

\paragraph{Overall pipeline.}
The continuity checker operates as an agent on top of the narrative KG and narrative stores.
Rather than a fixed sequence of operations, the agent retrieves candidate scene pairs, builds a
source-grounded evidence packet for each scene, performs LLM-based judgments, and updates the
continuity hypotheses.
Each cycle consists of three core behaviors:

\begin{itemize}[leftmargin=1.5em]
  \item \textbf{Candidate retrieval from the KG.}  
        Querying all $\mathcal{O}(N^2)$ scene pairs quickly becomes intractable as $N$ grows.  
        Instead, the agent first retrieves a sparse set of \emph{candidate} scene pairs from
        the scene--entity graph.  
        Two scenes are considered as candidates only if they share at least one canonical entity
        node or occasion-linked entity, such as a character, location, or object.
        This simple structural prior already yields a substantial reduction in search space:
        for example, in \textit{The Wandering Earth II}~\citep{gwo2023wanderingearth2} (173 scenes), naive all-pairs checking
        would require ${173 \times 172}/{2}$ pairwise decisions, whereas our entity-based
        filtering produces on the order of $1{,}500$--$2{,}000$ candidates.  
        The agent can further tighten this set by incorporating additional priors such as
        semantic similarity between scene summaries and constraints on script-order distance
        (e.g., only considering pairs whose scene indices differ by at most $k$).  

  \item \textbf{LLM-based assessment and hypothesis update.}  
        For each candidate pair, the agent invokes a structured continuity prompt conditioned on
        source-grounded scene packets: scene title/source span, linked characters, locations, objects,
        occasions, shared entities, and relevant events, interactions, or atomic facts.  The prompt
        asks whether the two scenes can share a production setup using only this evidence.
        The resulting binary decisions (with rationales) are used to update an evolving
        \emph{continuity graph} over scenes, where edges represent hypothesized continuity
        relations.

  \item \textbf{Reflective chain-level refinement.}  
        After accumulating pairwise judgments, the agent constructs preliminary continuity
        chains by extracting maximal cliques or connected components from the continuity graph.  
        It then invokes a second LLM prompt—the \emph{chain critic}—to reassess each chain
        holistically.  
        The chain critic examines the entire sequence, identifies weakly supported edges,
        over-extended clusters, or source-grounded contradictions, and proposes merges, splits, or
        drops that refine the chain structure.
\end{itemize}

This iterative loop—KG-based candidate retrieval, LLM pairwise evaluation, and LLM
chain-level reflection—constitutes the full continuity-checking agent.  The continuity-chain
visualization in Fig.~\ref{fig:continuity-visualization} is a presentation layer built on top of
these outputs rather than a step in the agent itself.

\paragraph{Pairwise continuity prompt.}
We use the following prompt to assess whether two scenes belong to the same production setup.

\begin{tcolorbox}[colback=SkyBlue!5,colframe=SkyBlue!100,halign=flush left,
title=Pairwise Production Continuity Judgment Prompt, breakable=true]
\small

\textbf{Role:} You are an expert reviewer evaluating whether two screenplay scenes can plausibly
share the same \emph{production setup}.

\textbf{Definition (Production Continuity):}\\
Two scenes are production-continuous if the provided source-grounded evidence supports reuse of a
compatible physical setting, recurring objects or set dressing, and stable character/setup
conditions.  Narrative time jumps are allowed, but do not infer visual continuity unless the evidence
supports it.  Treat missing production details as unknown rather than hallucinating costumes, props,
or lighting.

\vspace{4pt}

\textbf{Scene 1}\\
\textbf{Title/source:} \{title1, chunks1\}\\
\textbf{Evidence:} \{source\_evidence1\}\\
\textbf{Entities:} \{characters1, locations1, objects1, occasions1\}\\
\textbf{Facts:} \{events\_interactions\_facts1\}

\vspace{4pt}

\textbf{Scene 2}\\
\textbf{Title/source:} \{title2, chunks2\}\\
\textbf{Evidence:} \{source\_evidence2\}\\
\textbf{Entities:} \{characters2, locations2, objects2, occasions2\}\\
\textbf{Facts:} \{events\_interactions\_facts2\}

\vspace{6pt}
\textbf{Shared evidence:}\\
\{shared\_canonical\_evidence\}

\vspace{6pt}

\textbf{Decision Rules}
\begin{enumerate}[leftmargin=1.25em]
  \item \textbf{Use only provided evidence.} Do not invent production-design details.
  \item \textbf{Shared setup.} Favor continuity when scenes share a source-supported location,
        occasion, stable object, or recurring character configuration.
  \item \textbf{Blocking evidence.} Reject continuity when the source indicates incompatible
        locations, mutually exclusive scene functions, or a changed state that requires a different setup.
  \item \textbf{Uncertainty.} If evidence is sparse, lower confidence rather than fabricating a reason.
\end{enumerate}

\vspace{6pt}

\textbf{Return JSON fields:}
\begin{verbatim}
{
  "is_continuity": true | false,
  "confidence": 0.0-1.0,
  "support": ["evidence"],
  "blockers": ["blocker"],
  "reason": "brief source-grounded reason"
}
\end{verbatim}

\end{tcolorbox}

\paragraph{LLM chain-reflection prompt.}
After extracting initial chains, an agent-style reflection step evaluates whether the chain is
globally coherent and suggests refinements.

\begin{tcolorbox}[colback=SkyBlue!5,colframe=SkyBlue!100,halign=flush left,
title=Continuity Chain Critic Prompt, breakable=true]
\small

\textbf{Role:} You are a senior reviewer assessing an automatically grouped
\emph{continuity chain}.  Your task is to judge whether the chain is source-grounded and coherent as
a reusable production setup.

\textbf{Scenes:} \{ordered\_scene\_ids\} \\

\textbf{Scene packets:}\\
\{per\_scene\_source\_grounded\_packets\} \\

\textbf{Pairwise edges:}\\
\{pairwise\_edge\_decisions\} \\

\textbf{Evaluate:}
\begin{enumerate}[leftmargin=1.25em]
  \item Whether every neighboring scene pair has explicit source-grounded support.
  \item Whether the chain depends on a weak bridge scene that should be split.
  \item Whether shared entities, locations, objects, and occasions remain compatible across the chain.
  \item Whether any source-grounded fact contradicts reuse of the same setup.
  \item Whether the chain is redundant, over-extended, or should be merged with another chain.
\end{enumerate}

\textbf{Return JSON fields:}

\begin{verbatim}
{
  "coherence_score": 0.0-1.0,
  "confidence": 0.0-1.0,
  "decision": "keep" | "split" | "drop",
  "weak_edges": [
    ["scene_A", "scene_B"]
  ],
  "suggested_splits": [
    ["scene_A", "scene_B"],
    ["scene_C"]
  ],
  "rationale": "brief rationale"
}
\end{verbatim}
\end{tcolorbox}

\subsection{Character State Tracking}\label{app:character-status}

This component derives fine-grained, per-scene character states from scene-structured narrative text.  
Whereas the main framework focuses on entities, relations, and event-centric structure,
the character-state module targets a different
layer of narrative grounding: short, observable descriptions of what important
characters are doing in each scene.

\paragraph{Iterative extraction with reflection.}
Following the reflection-driven framework used across our system, the agent performs a
recurrent \emph{extract $\rightarrow$ reflect $\rightarrow$ revise} cycle.  
For each scene, the system (1) retrieves candidate characters from the scene--entity graph,  
(2) extracts only \emph{visible, non-speculative} states using a structured LLM prompt, and  
(3) applies a critic model that checks coverage, grounding, and precision.  
Feedback is reinjected into the next extraction pass until convergence or a retry limit is
reached.  
This loop yields a stable, scene-aligned collection of character states across the entire screenplay.

\paragraph{Observable state representation.}
Each character state captures only what is externally observable—actions, expressions,
posture, and tone—excluding psychological or symbolic interpretation.  
Outputs use a minimal structured representation listing only important characters for each scene.

\paragraph{Visualization.}
To support interactive exploration, we render the extracted dataset using a timeline swimlane view.

\subparagraph{1.\ Timeline Swimlane View.}
A global \emph{character--scene matrix} showing all characters (rows) across all scenes
(columns).  
This view offers a high-level visualization of appearance frequency, co-occurrence
structure, and participation patterns, enabling users to quickly identify which characters
move together across the narrative.  In a production setting, the same view can support assistant
directors, continuity supervisors, and wardrobe/makeup teams by exposing character availability,
scene grouping, co-presence, and state-continuity constraints across the shooting plan.

\begin{figure*}[t]
  \centering
  \includegraphics[width=\linewidth]{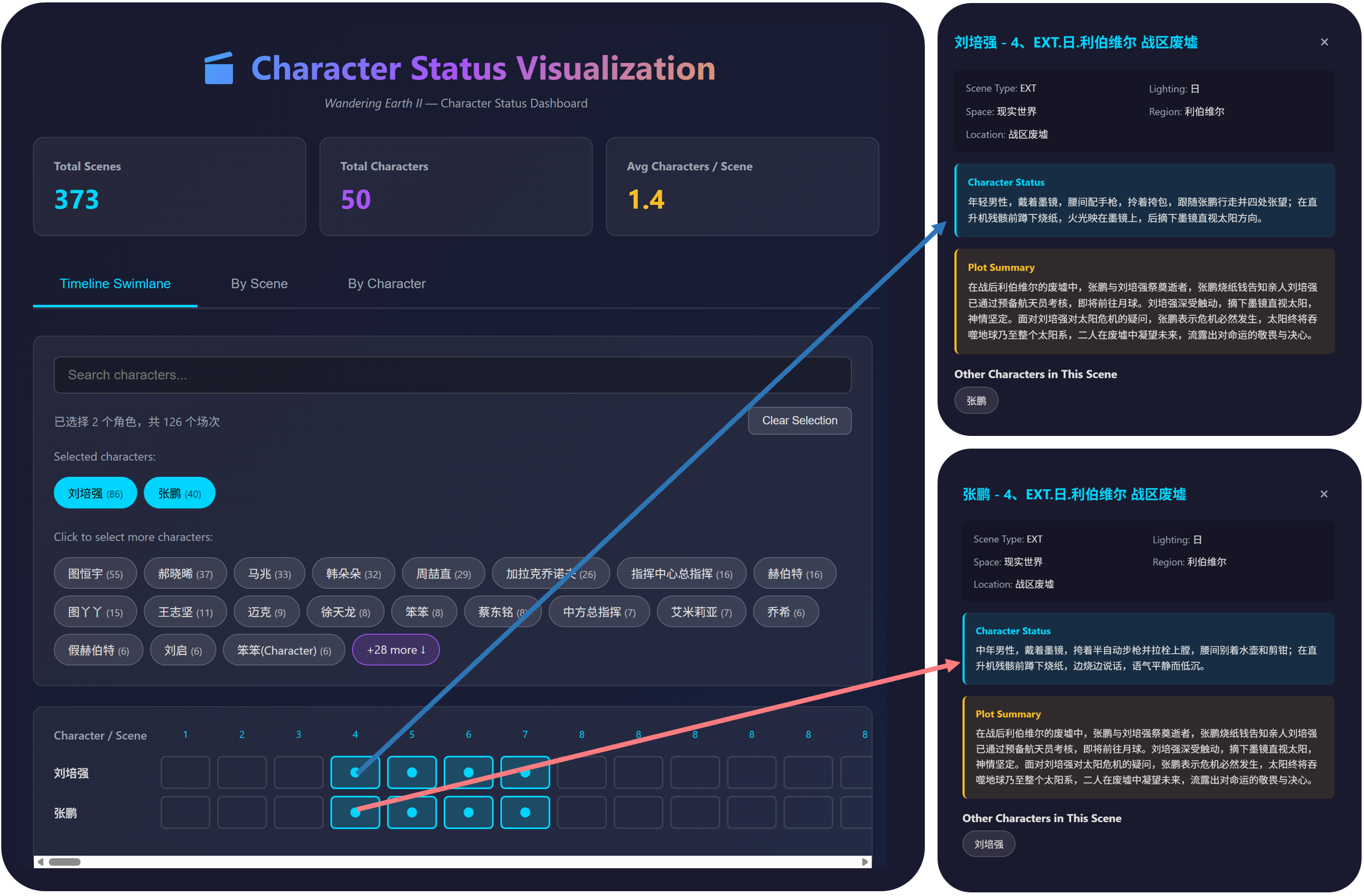}
  \caption{\textbf{Timeline Swimlane View.}
  A screenplay-wide character--scene matrix enabling global inspection of appearance patterns.}\label{fig:swimlane-view}
\end{figure*}

\end{document}

%% file: math_commands.tex
\usepackage{amsmath,amsfonts,bm}

\def\eqref#1{equation~\ref{#1}}

\def\1{\bm{1}}

\DeclareMathAlphabet{\mathsfit}{\encodingdefault}{\sfdefault}{m}{sl}
\SetMathAlphabet{\mathsfit}{bold}{\encodingdefault}{\sfdefault}{bx}{n}

